%% file: main.tex
\documentclass[10pt,journal,compsoc]{IEEEtran}

\ifCLASSOPTIONcompsoc
  \usepackage[nocompress]{cite}
\else
  \usepackage{cite}
\fi

\ifCLASSINFOpdf
\else
\fi

\usepackage{cite}
\usepackage{amsmath,amssymb,amsfonts}
\usepackage{algorithmic}
\usepackage{graphicx}
\usepackage{textcomp}
\usepackage{xcolor}
\usepackage{hyperref}
\usepackage{amsmath}
\usepackage{xparse}
\usepackage{pgf}
\usepackage{pdfpages}

\begin{document}
\title{Unsupervised Domain-agnostic Fake News
Detection using Multi-modal Weak Signals}
\author{Amila Silva, Ling Luo, Shanika Karunasekera, and Christopher Leckie% <-this % stops a space
\IEEEcompsocitemizethanks{\IEEEcompsocthanksitem Amila Silva, Ling Luo, Shanika Karunasekera, and Christopher Leckie are with the School of Computing and Information Systems, The University of Melbourne, Australia, VIC, 3010. 
E-mail: \{amila.silva@student., ling.luo@, karus@, caleckie@\}unimelb.edu.au}}% <-this % stops a space

%{Silva \MakeLowercase{\textit{et al.}}: Unsupervised Domain-agnostic Fake News Detection using Multi-modal Weak Signals}

\IEEEtitleabstractindextext{%
\begin{abstract}
The emergence of social media as one of the main platforms for people to access news has enabled the wide dissemination of fake news, having serious impacts on society. Thus, it is really important to identify fake news with high confidence in a timely manner, which is not feasible using manual analysis. This has motivated numerous studies on automating fake news detection. Most of these approaches are supervised, which requires extensive time and labour to build a labelled dataset. Although there have been limited attempts at unsupervised fake news detection, their performance suffers due to not exploiting the knowledge from various modalities related to news records and due to the presence of various latent biases in the existing news datasets (e.g., unrealistic real and fake news distributions). To address these limitations, this work proposes an effective framework for unsupervised fake news detection, which first embeds the knowledge available in four modalities (i.e., source credibility, textual content, propagation speed, and user credibility) in news records and then proposes $(UMD)^2$, a novel noise-robust self-supervised learning technique, to identify the veracity of news records from the multi-modal embeddings. Also, we propose a novel technique to construct news datasets minimizing the latent biases in existing news datasets. Following the proposed approach for dataset construction, we produce a Large-scale Unlabelled News Dataset consisting 419,351 news articles related to COVID-19, acronymed as \textsc{LUND-COVID}. We trained the proposed unsupervised framework using \textsc{LUND-COVID} to exploit the potential of large datasets, and evaluate it using a set of existing labelled datasets. Our results show that the proposed unsupervised framework largely outperforms existing unsupervised baselines for different tasks such as multi-modal fake news detection, fake news early detection and few-shot fake news detection, while yielding notable improvements for unseen domains during training.
\end{abstract}

}

\maketitle

% is not selected <OR> if conference mode is selected - because compsoc
% conference papers position the abstract like regular (non-compsoc)
\IEEEdisplaynontitleabstractindextext

% \fi
\IEEEpeerreviewmaketitle

\section{Introduction}\label{sec:introduction}
\textbf{Motivation. }With the increasing accessibility of the Internet and the ease of social media usage, fake news has become a significant social problem. The damage caused by fake news is particularly significant during major events such as presidential elections and the COVID-19 pandemic. For example, in~\cite{islam2020covid}, it has been estimated that at least 800 people died and 5800 were admitted to hospital as a result of false information related to the COVID-19 pandemic, e.g., believing alcohol-based cleaning products are a cure for the virus. The study in~\cite{nieves2021infodemic} also supports this claim by showing that COVID-19 mortality rates are relatively higher for countries with heavy use of social networks for obtaining information. Hence, the early detection is essential to stop spreading fake content during emergencies. Identifying fake news manually is not practical due to the high volume of news items that circulate on a daily basis. Thus, building data-driven automated solutions for fake news detection~\cite{shu_fakenewsnet_2018,pierri_false_2019,zhou_fake_2018} has attracted considerable research effort. Nevertheless, the performance of most existing fake news detection techniques~\cite{shu_hierarchical_2019,shu_leveraging_2020,wang_eann_2018,monti_fake_2019,shu_defend_2019}, which are mostly based on supervised learning techniques, largely relies on the availability of large-scale labelled datasets. Since compiling clean labels for such large datasets is extremely expensive and time-consuming, such supervised fake news detection techniques are not ideal to stop the spread of fake content during emergencies. \textbf{\textit{This work proposes a novel unsupervised learning framework to detect fake news during emergencies using multiple weak signals from different modalities related to fake news}}. Although there are a few previous attempts on unsupervised fake news detection~\cite{yang_unsupervised_2019,hosseinimotlagh2018unsupervised,gangireddy2020unsupervised,li2021unsupervised}, our work has several key advantages over these works due to the following important research contributions. 

First, the theories motivated from social science and empirical observations show that various informative modalities of news articles (e.g., the psycholinguistic features of the text modality, the credibility of the news source and social engagements) could be potentially useful to identify the veracity of fake news. Most of the previous unsupervised fake news detection models based on such modalities are either not scalable or not generalizable to unseen domains during training. To address this gap, \textbf{\textit{we propose domain-agnostic pre-training objective functions for different modalities in news articles -- e.g., source credibility, textual content, propagation speed and social engagements, using which the information embedded in each modality can be encoded using large-scale news datasets.}} Subsequently, such embeddings can serve as features for downstream multi-modal fake news detection models.  

Second, a few recent supervised approaches have shown that exploiting multiple modalities~\cite{silva2021embracing,shu_leveraging_2020} together could boost the prediction accuracy of fake news. Nevertheless, almost all existing unsupervised fake news detection techniques are restricted to one modality (See Section~\ref{sec:related_work} for a detailed comparison between our work and these unsupervised frameworks). Thus, \textbf{\textit{this work devises $(UMD)^2$, a novel self-supervised learning technique, to effectively exploit the embeddings from multiple sources to yield improved performance for unsupervised fake news detection.}} Since some news articles may not have all modalities available, $(UMD)^2$ has been designed to handle the partial availability of modalities. 

Third, although there are many well-known datasets on fake news detection~\cite{shu_fakenewsnet_2018,cui2020coaid,zhou2020recovery}, a recent work~\cite{zhou2021hidden} has shown that the widely-used data collection strategies for news datasets could introduce various selection biases that hinder the generalizability of the models learned using such datasets. For example, most existing datasets~\cite{shu_fakenewsnet_2018,zhou2020recovery} do not reflect the empirical distribution of fake and real news in more realistic settings (see Section~\ref{sec:news_dataset} for more details). Such biases could mislead machine learning models, especially when they are learned without a strong supervisory signal. Also, almost all previous unsupervised fake news detection models~\cite{yang_unsupervised_2019,li2021unsupervised} have been trained and evaluated using relatively small datasets despite the true potential of unsupervised frameworks being typically reliant on using large-scale datasets. To bridge these gaps, \textbf{\textit{we propose a novel data collection strategy to produce large-scale news datasets while minimizing various selection biases in the existing datasets.}} Following this technique, \textbf{\textit{we produce \textsc{LUND-COVID}, a Large-scale multi-modal Unlabelled News Dataset on COVID-19 covering more than 400,000 news articles, and make the dataset publicly available}}, which could be a valuable resource for future research on unsupervised fake news detection.    

\textbf{Contribution. }Our contributions in this work can be summarized as follows:
\begin{enumerate}
    \item We propose domain-agnostic pre-training objectives to encode the knowledge available in four different modalities in a news article: the source of the articles; the textual content; the propagation pattern; and the engaging users with the article. We qualitatively show their varying importance for fake news detection.
    \item We devise $(UMD)^2$, a self-supervised multi-modal fake news detection technique, which adopts a teacher-student architecture to detect fake news articles using their multi-modal embeddings. $(UMD)^2$ is designed in a manner to be robust against the partially available noisy modalities with varying importance.
    \item We propose \textit{backward news dataset construction}, a novel data collection strategy to produce large-scale news datasets while minimizing various selection biases in the existing datasets. Using the proposed data collection approach, we produce \textsc{LUND-COVID}, a \textbf{L}arge-scale multi-modal \textbf{U}nlabelled \textbf{N}ews \textbf{D}ataset on \textbf{COVID}-19. We quantitatively show that the proposed data collection approach can produce a realistic dataset with respect to the empirical distribution of real and fake news records and media outlet coverage. 
    \item We train the proposed modality-specific pre-training objectives and $(UMD)^2$ using \textsc{LUND-COVID}, and evaluate them using a set of existing labelled datasets. We evaluate $(UMD)^2$ under three settings, namely: multi-modal fake news detection; fake news early detection; and few-shot fake news detection. Our quantitative evaluation shows that $(UMD)^2$ outperforms existing state-of-the-art unsupervised fake news detection methods by as much as 12\% in F1-score. 
\end{enumerate}

\section{Problem Statement}\label{sec:problem_statement}

Let $R$ be a set of news records. Each record $r \in R$ is represented as a tuple $\langle s^r, t^r, p^r, u^r\rangle$ consisting of different modalities in $r$'s lifespan -- from its origin from a media outlet until either it is deleted or fully propagated on social media, that are informative to identify $r$'s veracity. In this representation, (1) $s^r$ is the media outlet from which $r$ was produced (i.e., the domain of the URL of $r$); (2) $t^r$ represents the textual content of $r$ (e.g., article title and body text); (3) $p^r$ is the propagation speed of $r$ across social media platforms at different time-steps as a time-series; and (4) $u^r$ represents the characteristics of the users (e.g., the number of followers, the verification status and the age of the user profile) engaging with $r$ in social media. We elaborate these modalities in detail in Section~\ref{sec:embedding}. Our unsupervised fake news detection problem aims to learn a mapping function between the veracity labels $y^r$ and the modalities $\langle s^r, t^r, p^r, u^r\rangle$ of news records $r \in R$ without having a labelled dataset. This consists of two subtasks:
\begin{enumerate}
    \item Subtask I aims to learn in an unsupervised manner a representation of the knowledge in each modality in $\langle s^r, t^r, p^r, u^r\rangle$ as a low-dimensional vector $\langle z^s, z^t, z^p, z^u\rangle$. This learning task uses the knowledge motivated by various computational social theories or previous empirical studies, such that each modality's ability to identify the veracity of $r$ is preserved -- e.g., this task learns $f^u:u^r \rightarrow z^u$, where $z^u\in \mathbb{R}^d$ and $f^u$ is the unsupervised embedding function, to represent the knowledge in $u^r$.
    \item Subtask II learns a self-supervised mapping function $g$ to produce the veracity label $y^r$ of 1 if $r$ is fake, or 0 otherwise, using $r$'s embeddings either from all the modalities $\langle z^s, z^t, z^p, z^u\rangle$ or a set of selected modalities $\langle z^s, z^t\rangle$ -- i.e., $g:\langle z^s, z^t, z^p, z^u\rangle \rightarrow y^r$.
\end{enumerate}

We describe the proposed solutions for Subtasks I and II in Sections~\ref{sec:embedding} and~\ref{sec:inference}, respectively.

\section{Embedding Weak Sources}\label{sec:embedding}
% \begin{figure}
%     \centering
%     \includegraphics[width=\linewidth]{news_lifespan.pdf}
%     \caption{The lifespan of a news record: a news record is generated from media outlet and it is then propagated to users mostly via today's social networks, where the users interconnected with each other (dashed lines).}
%     \label{fig:lifespan}
%     \vspace{-5mm}
% \end{figure}

This section describes the proposed unsupervised mapping functions that encode the knowledge in the selected weak signals: source credibility ($s^r$), textual content ($t^r$), propagation speed ($p^r$), and user credibility ($u^r$), all related to misinformation\footnote{This manuscript uses the terms misinformation and fake news interchangeably, referring to the misinformation definition in~\cite{zhou_fake_2018} -- false information that is spread, regardless of whether there is intent to mislead.} and motivated by either computational social theories~\cite{zhou_fake_2018} or empirical studies in previous works~\cite{shu_fakenewsnet_2018,shu_leveraging_2020,zhou2020recovery}. We select weak signals for this study as they cover various stages of the lifespan of a news record. % (see Fig.~\ref{fig:lifespan}). 
We provide strong justifications for the selection of these signals later.

To exploit weak signals for misinformation detection, most previous works~\cite{shu_leveraging_2020,zhou2020recovery} adopt different statistical measures to convert such signals as hard weak labels for fake news labels. For example, the study in~\cite{shu_leveraging_2020} computes the credibility of a user as a score, and if a news record has an average user credibility score less than a particular threshold, then the news record is labelled as fake, or otherwise as real. The knowledge in weak signals that we can represent using such hard labels could be limited. 

To address this challenge, we propose embedding functions to represent the knowledge of a news record related to each weak signal as a low-dimensional vector. We elaborate the proposed embedding functions in this section. Since such embedding functions typically require very large datasets for effectively training, we adopt \textsc{LUND-COVID}, a large-scale unlabelled news dataset on COVID-19 that we have processed and released in this work. We present more details about this dataset in Section~\ref{sec:news_dataset}. We adopt three publicly available labelled datasets: PolitiFac~\cite{shu_fakenewsnet_2018}; GossipCop~\cite{shu_fakenewsnet_2018}; and CoAID~\cite{cui2020coaid}, to qualitatively analyse the performance of the proposed pre-training objectives. Table~\ref{tab:statistics} presents the descriptive statistics of these datasets.

\begin{table}[t]
\scriptsize
    \centering
    \begin{tabular}{c|c|c|c|c}
        & Train & \multicolumn{1}{c|}{In-Domain Test} & \multicolumn{2}{c}{Out-of-Domain Test} \\
        \hline
        & LUND-COVID& CoAID &PolitiFact & GossipCop \\
        \hline\hline
        $\#\; articles$ &  419,351& 2,869&499& 3,735\\
        \hline
        $\#\; tweets$ &  17,802,652& 151,964& 724,964& 1,157,795\\
        \hline
        $\frac{\#\; tweets}{\#\; articles}$ &42.5 & 52.9 & 1452.8& 310.0\\
    \end{tabular}
    \caption{Descriptive statistics of the collected unlabeled training dataset and the selected labelled test datasets}
    \label{tab:statistics}
    \vspace{-5mm}
\end{table}

\subsection{Source Credibility}\label{subsec:source_credibility}
Several recent works~\cite{nakov2020can,pehlivanoglu2021role} have shown that the credibility or trustworthiness of media outlets can be used as a weak signal to fact-check the news articles that are published by the corresponding media outlets. These works attempt to assign a credibility score for each news outlet based on its previous publications and propagate the same label for future publications coming from the same outlet. Following these attempts, there are publicly available databases from which the different media outlets can be labelled as either reliable ($s^r=1$) or unreliable ($s^r=0$). However, this approach has two limitations: (1) there could be media outlets that post a mixture of real and fake news, thus, hard source credibility score may not be suitable for such an outlet; (2) there could be new or little-known media outlets that are not covered by existing databases to identify their source credibility. Most of these little-known media outlets are the alternative media attached to main media outlets. As found by~\cite{horne2019different,horne2018exploration}, most mainstream outlets publish their news article on the mainstream with or without slight modification via various alternative media outlets to maximize the reach of their articles. We exploit this content sharing behaviour to address the aforementioned limitations.  

\begin{figure}[t]
    \begin{center}
    \resizebox{\linewidth}{!}{\input{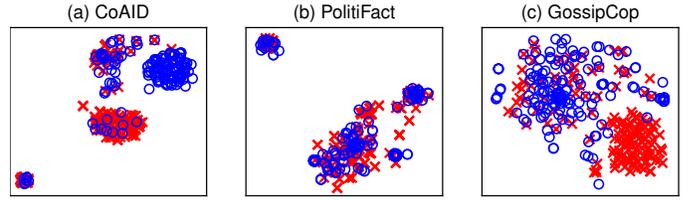}}
    \end{center}
    \vspace{-5mm}
    \caption{t-SNE visualization of the source-based embedddings of the news articles in CoAID, PolitiFact, and GossipCop. Fake and real news records are plotted using red crosses and blue circles respectively.}
    \label{fig:source_embeddings}
    \vspace{-5mm}
\end{figure}

Our approach can be elaborated as follows:
\begin{enumerate}
    \item We first employ an existing database $D^s$ proposed in~\cite{zhou2020recovery,baly:2018:EMNLP2018,baly:2020:ACL2020} to construct a network $G^d$ of news sources (i.e., URL domains) -- $G^d$ has a node for each unique media-outlet in $D^s$ and connects the media-outlets that have the same credibility label (reliable, unreliable or mixed) in $D^s$.
    \item Following~\cite{horne2019different,horne2018exploration}, we construct a network of news articles $G^r$ to represent the aforementioned news content sharing behavior: compute the pairwise cosine similarity between TF-IDF features of the news articles computed using their textual content (i.e., title and body text); and connect article pairs with a cosine similarity of 0.85 or above (as suggested by~\cite{horne2019different,horne2018exploration}) in our network. The intuition is that the articles related to the same mainstream outlet should be connected in $G^r$. 
    \item We combine $G^d$ and $G^r$ into a graph $G^s$ by connecting the articles in $G^r$ to the corresponding media-outlet in $G^d$. If the media-outlet of an article cannot be found in $G^d$, we add an isolated node to $G^d$.
    \item We learn representations for the nodes in $G^s$ such that the connected nodes have similar representations~\cite{silva2020meteor,silva2020omba,zhang2017react}. For a given node $n$, its embedding $z^n$ is updated by minimizing the following:
    \begin{equation}\label{eq:source_embeddings}
        L = -log(\sigma (z^n \cdot z^{n^{+}})) - log(\sigma (-z^n \cdot z^{n^{-}}))
    \end{equation}
    where $n^{+}$ $(n^{-})$ denotes a connected (disconnected) node to $n$ in $G^s$. $\sigma()$ is the sigmoid function.
    \item After training the embeddings using Eq.~\ref{eq:source_embeddings}, the embedding of a node corresponding to an article $r$ is returned as its source-based representation $z^r$.  
\end{enumerate}

We jointly learn source-based representations for the articles in LUND-COVID, CoAID, PolitiFact, and GossipCop using the approach proposed above. We empirically observed that having a very large dataset like LUND-COVID helps to make $G^s$ dense and to propagate information via the edges of $G^s$ effectively, especially for the articles whose media-outlets are not covered by $D^s$. To check the informativeness of the learned embeddings for misinformation detection, Figure~\ref{fig:source_embeddings} adopts t-SNE~\cite{maaten2008visualizing} to visualize the learned embeddings for the articles in the labeled datasets. As can be seen, the source-based embeddings are considerably different for fake and real news. Thus, we can observe different clusters for real and fake news in the embeddings space, which verifies the informativeness of the source modality for misinformation detection. However, the clusters are not perfect. This is why it is important to exploit multimodalities together when identifying fake news in an unsupervised manner instead of relying on one signal~\cite{zhou2020recovery}.    

\subsection{News Content}\label{sec:news_content}
Many previous works on fake news detection show that there are significant differences between the textual content of real news and fake news. Some of these works~\cite{silva2021embracing} exploit the semantical and syntactical features of the textual content. However, these works do not generalize well for unseen media-outlets during training due to the significant semantic and syntactic differences between media-outlets. 

As a solution, this work focuses on high-level affective features of the textual content such as psycholinguistic features, sentiment-specific features and moral features, as these features are typically consistent across different media-outlets. Specifically, we extract 111 high-level features for each new record using its title and body text as its textual content, that belongs to six categories: (1) \textit{Sentiment} score from the fine-tuned roBERTa~\cite{liu2019roberta} for sentiment classification, (2) \textit{Emotions} features from the NRC emotions lexicon~\cite{mohammad2013nrc}, (3) \textit{Psycholinguistic} features from the LIWC dictionary~\cite{pennebaker_development_2015}, (4) \textit{Readability} measure using the SMOG metric~\cite{mc1969smog}, (5) \textit{Morality} features from the Moral Foundations Dictionary~\cite{graham2009liberals}, and (6) \textit{Hyperbolic} features from the dictionary proposed in~\cite{chakraborty2016stop} (see our supplementary material for more details about these features). The feature vector is then normalized and compressed using an autoencoder $f^t$ with the clustering-specific unsupervised objective function proposed in~\cite{dahal2018learning}. This objective function adopts an autoencoder to compress the textual-content based features to produce its text-based embedding using its encoder -- i.e., $f^t:t^r \rightarrow z^t$, such that $z^t$ can be used to reconstruct the original textual-content based features using the decoder of the autoencoder and the embedding space of $z^t$ has two clear clusters. The latent representation of this autoencoder (i.e., the output of the encoder) for a news $r$ is used as the textual representation $z^t$ of $r$. %Please see~\cite{dahal2018learning} for more details about the objective function.

\begin{figure}
    \begin{center}
    \resizebox{\linewidth}{!}{\input{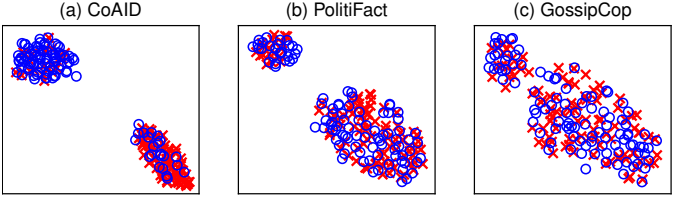}}
    \end{center}
    \vspace{-5mm}
    \caption{t-SNE visualization of the text-based embedddings of the news articles in CoAID, PolitiFact, and GossipCop. Fake and real news records are plotted using red crosses and blue circles respectively.}
    \label{fig:text_embeddings}
    \vspace{-5mm}
\end{figure}

After training $f^t$ using \textsc{LUND-COVID}, Figure~\ref{fig:text_embeddings} shows the produced text-based embeddings for the labelled datasets from $f^t$. Due to the clustering-specific objective function used to train $f^t$, the pre-trained text-based embedding spaces show clear clusters despite being noisy with respect to the true veracity labels. 

\subsection{Propagation Speed}\label{sec:propagation_speed}

Fake news are typically written in a way to maximize their reach over social media networks such as Twitter. As a result, it has been found in previous studies~\cite{shu_fakenewsnet_2018} that fake news propagates faster compared to real news. As shown in~\cite{shu_fakenewsnet_2018}, there is also a sudden increase in the number of tweets/retweets for fake news. In contrast, real news mostly shows a steady increase in the number of tweets/retweets. The sudden increase for fake news may not appear just after the news is posted. Here, we propose an embedding technique to represent these observations related to the propagation of a news record as a low-dimensional vector.
% \begin{figure}
%     \centering
%     \includegraphics[width=\linewidth]{temporal_clustering.pdf}
%     \caption{Overview of the embedding technique for propagation speed}
%     \label{fig:propagation_speed}
% \end{figure}
\begin{figure}
    \begin{center}
    \resizebox{\linewidth}{!}{\input{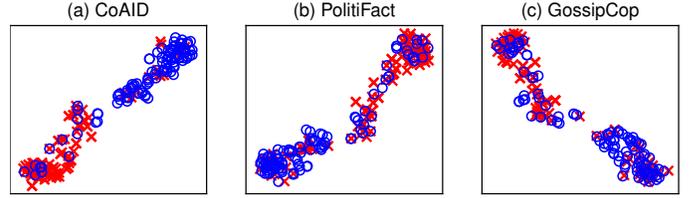}}
    \end{center}
    \vspace{-5mm}
    \caption{t-SNE visualization of the propagation-based embedddings of the news articles in CoAID, PolitiFact, and GossipCop. Fake and real news records are plotted using red crosses and blue circles respectively.}
    \label{fig:propagation_embeddings}
    \vspace{-5mm}
\end{figure}

Our approach can be elaborated as follows. 
\begin{enumerate}
    \item For each news record $r$, we represent its propagation as a time-series $p^r=\{p_0^r, p_1^r, ..., p_T^r\}$, where $p_t^r$ represents the number of tweets/retweets of $r$ that are posted between $t\Delta-(t+1)\Delta$, $t\in\{0, 1, ..., T\}$. We set $\Delta=1\;hour$ and $T=48$ in this study.
    \item Here, the goal is to learn the mapping function $f^{p}: p^r \rightarrow z^p$, where $z^p \in \mathbb{R}^d$ such that the learned representations are able to capture the scale and the sudden increases in $p^r$, while being invariant to the time-point where such sudden increases occur. To learn $z^p$, we adopt the contrastive learning approach proposed in~\cite{eldele2021time} that learns a representation for time-series data by maximizing the similarity among different correlated views of the same sample, while minimizing their similarity with other samples. The knowledge that should be preserved from such contrastive learning can be controlled by having appropriate data augmentation techniques to generate correlated views. Following~\cite{eldele2021time}, we adopt a weak and a strong augmentation technique to generate correlated views. For the weak augmentation, we add random noise to the time series -- i.e., $p^r_{weak} = \{p_0^r + \delta_0^r, p_1^r + \delta_1^r, ..., p_T^r+ \delta_T^r\}$. To make our representations invariant to the locations of sudden increases, a permutation-and-jitter strategy is used as strong augmentation. Specifically, we split $p^r$ into a random number of segments and randomly shuffle them and noise to the permuted signal -- i.e.,  $p^r_{strong} = \{p_8^r + \delta_0^r, p_9^r + \delta_1^r, ..., p_3^r+ \delta_T^r\}$.
    \item Following~\cite{eldele2021time}, we formulate $f^p$ using a LSTM and a multi-head attention-based transformer architecture to produce a low-dimensional representation from $p^r$. The trainable parameters in $f^p$ are learned using two objectives: (1) temporal contrasting loss terms to preserve the properties of the time-series by predicting the intermediate latent representation (output from the LSTM) at $t+1$ using the hidden representations seen up to that timestep; (2) contexual contrasting loss term to preserve the similarity of the correlated views by maximizing the dot product between the representation from augmented views of the same instance -- i.e., $f^p(p^r_{strong}) \cdot f^p(p^r_{weak})$, and minimizing the similarity of the views from different instances -- i.e., $f^p(p^r_{strong}) \cdot f^p(p^{r^{'}}_{weak})$. Please see~\cite{eldele2021time} for more details.
\end{enumerate} 

We train $f^p$ using the news records in \textsc{LUND-COVID}. Figure~\ref{fig:propagation_embeddings} visualizes the produced embedding for the labelled datasets using $f^p$, which shows the differences in the propagation of real and fake news records, especially for CoAID. Although $f^p$ is trained using COVID-19 related articles, the pattern in the embedding space is quite consistent for the out-of-domain datasets too. Such domain-agnostic feature spaces will assist the downstream fake news detection model to generalize well for the out-of-domain datasets.

\subsection{User Credibility}

\begin{figure}
    \begin{center}
    \resizebox{\linewidth}{!}{\input{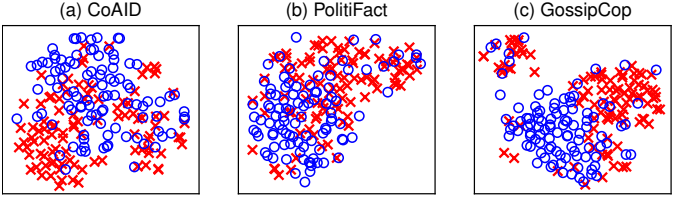}}
    \end{center}
    \vspace{-5mm}
    \caption{t-SNE visualization of the user-based embedddings of the news articles in CoAID, PolitiFact, and GossipCop. Fake and real news records are plotted using red crosses and blue circles respectively.}
    \label{fig:user_embeddings}
    \vspace{-5mm}
\end{figure}

The credibility of the users engaging with a particular news record has been identified as a strong signal to identify the veracity of news records~\cite{silva2021propagation2vec,silva-embedding-2020,shu_defend_2019}. Motivated by this finding, here we propose a pre-training objective function to learn a representation $z^u$ for a given news record $r$ using its user engagements $u^r$ such that $z^u$ can preserve the credibility of the engaging users with $r$. 

In our approach, the knowledge in the social engagements ($u^r$) of a news record $r$ is initially converted into a network structure (i.e.,  \textit{engagement-based graph} and $G^u$), denoted using a tuple -- $G^u = (V^u, A^u, X^u)$, where $V^u$ represents the set of users who engaged with $r$ and there is a special node $v_\star^{u}$ in $V^u$ to represent the news article itself. $A^u$ is the adjacency matrix representing user-user and user-news connections based on uses' retweeting patterns. $X^u$ represents the features of $V^u$ to characterize different users. We adopt the approach proposed in~\cite{silva-embedding-2020} to construct $G^u$. %Please refer to~\cite{silva-embedding-2020} for more details. %or the supplementary material for more details. 

Based on the knowledge in the engagement-based network of a record $G^u$, here we propose a mapping function $f^u: G^u \rightarrow z^u$ that learns a low-dimensional representation for the corresponding news record in a self-supervised manner. Since the number of nodes in $G^u$ is not fixed and could be different in size, we adopt a Graph Attention Network (GAT) architecture~\cite{velivckovic2017graph} to model $f^u$. $f^u$ is learned using a objective function that is motivated by Deep Graph Infomax~\cite{velickovic2019deep}. Deep Graph Infomax relies on maximizing mutual information between patch representations and corresponding high-level summaries of graphs that are derived from $f^u$. Since Deep Graph Infomax was originally proposed to learn representations for the nodes in a graph or to summarize the nodes' representation, we adopt a modified learning procedure as follows to learn representation for a news record $r$ using its $G^u = (V^u, A^u, X^u)$:
\begin{enumerate}
    \item Step 1: Randomly sample a set of nodes from $V^u\setminus\{v_\star^u\}$ and invert their feature value to produce a negative example $\tilde{G}^u = (V^u, A^u, \tilde{X}^u)\sim \mathbb{C}(V^u, A^u, X^u)$ of $G^u$. Here, we adopt value inversion as the corruption function as most of the selected user features in $X^u$ have been shown to be monotonically increasing/decreasing with the veracity of the news articles. 
    \item Step 2: Obtain the node representations for the nodes in $G^u$ by passing them through the encoder: $z^u = f^u(G^u) = \{z_\star^u, z_1^u, z_2^u, ...\}$
    \item Step 3: Obtain the node representations for the nodes in $\tilde{G}^u$ by passing them through the encoder: $\tilde{z^u} = f^u(\tilde{G}^e) = \{\tilde{z}_\star^u, \tilde{z}_1^u, \tilde{z}_2^u, ...\}$
    \item Step 4: As the summarized representation, we select the representation of the node corresponding to the news article from Step 2 -- i.e., $z_\star^u$
    \item Step 5: Update the parameters in $f^u$  and $d^u$ by maximizing the following equation:
\begin{equation*}\label{eq:infomax}
\frac{\sum_{v_i \in V^u\\\{v_\star^u\}} \log (d^u(z_i^u, z_\star^u)) + \log (1-d^u(\tilde{z}_i^u, z_\star^u))}{|V^u|-2}
\end{equation*}
where $d^u$ is a discriminator that assigns higher probability scores for positive node and summary pairs, and lower scores for negative pairs.
    \item Step 6: Return $z_\star^u$ as the embedding of $r$.
\end{enumerate}

We apply Steps 1-6 to constructing the engagement-based graphs for the records of \textsc{LUND-COVID} to pre-train $f^u$. Subsequently, for a given news record $r$ in our labelled datasets, its $f^u$ is used as the user credibility-based embedding $z^u$, which are visualized in Figure~\ref{fig:user_embeddings}. As can be seen, the user-based embedding spaces do not show a clear separation between fake and real news as in most other modalities such as source credibility and propagation speed. This observation motivates the importance of assigning varying importance to different modalities when exploiting multi-modalities for fake news detection.

\section{$(UMD)^2$: Unsupervised Misinformation Detection Using Multi-modal Data}\label{sec:inference}
\subsection{Overview}
The goal of $(UMD)^2$ is to find the veracity label $y$ of a news record $r$ in an unsupervised manner given its embeddings from multi-modalities -- e.g., source credibility-based embedding $z^s$; textual content-based embedding $z^t$; propagation speed-based embedding $z^p$; and user credibility-based embedding $z^u$. $(UMD)^2$ is a general technique which can easily scale to other modalities or multi-modal applications. 

$(UMD)^2$ can identify the importance of different modalities in each news record and aggregate the corresponding multi-modal embeddings accordingly with the help of a modified gated multi-modal unit (GMU) model. Despite being able to recover the informativeness of modalities in a data-driven manner, the conventional GMU~\cite{ovalle2017gated} is \textbf{\textit{unable to exploit the explicitly known information about the uninformative/missing modalities}} for some instances or applications. For example, the social engagements may not be sufficient and informative for fake news early detection. In Section~\ref{subsec:multi-modal-agg}, we propose a modified GMU to exploit such explicitly known information, which could potentially improve the applicability of $(UMD)^2$ under different settings.        

$(UMD)^2$ also handles the challenge of \textbf{\textit{having partially available modalities}} using a teacher-student architecture. Both teacher and student networks in $(UMD)^2$ share the same architecture. However, the teacher network will have access to all the modalities of a training instance (i.e., an unmasked training instance) while the student network attempting to mimic the operation of the teacher network using a masked version (i.e., after removing a set of modalities deliberately) of the same instance. The main objective of this architecture is to produce the performance of a model trained using a complete dataset -- i.e., teacher network, using a model from a partiality available dataset -- i.e., student network, which is shown to be promising to handle missing values in the literature~\cite{baevski2022data2vec,silva2021propagation2vec,grill2020bootstrap}. Section~\ref{subsec:teacher-student} provides detailed information about this architecture. 

\textbf{\textit{The multi-modal embeddings from pre-trained tasks are typically noisy}} when applying them to address various downstream tasks in an unsupervised manner. Several noise-robust learning techniques~\cite{liu2020peer,berthon2021confidence,silva2022noise} are proposed in the literature to handle such noise. Motivated by these works, we propose a pseudo-labelling based noise-robust loss function for unsupervised learning to train $(UMD)^2$. As detailed in Section~\ref{subsec:noise-robust-misinformation-detection}, the proposed noise-robust loss function adopts the teacher model $(UMD)^2$ to produce the pseudo labels, which are subsequently used to learn the student model. The following sections provide detailed information about the aforementioned contributions in $(UMD)^2$. 

\subsection{Modified GMU for Multi-modal Aggregation}\label{subsec:multi-modal-agg}

% \begin{figure}
%     \centering
%     \includegraphics[width=0.7\linewidth]{GMU_Model.pdf}
%     \caption{Overview of the modified GMU module, where the attention weights for each modality in news record $r$ are decided by their uni-modal representations $\{z^s, z^t, z^p, z^u\}$ and the explicit informativeness scores of the modalities $m$.}
%     \label{fig:gmu_model}
%     \vspace{-5mm}
% \end{figure}

This section presents detailed information about the proposed variant of the Gated Multimodal Unit to assign varying importance for different modalities based on their informativeness while exploiting the explicit information about missing modalities. For given pre-trained multi-modal embeddings $\{z^s, z^t, z^p, z^u\}$ of a news record $r$ and the explicit knowledge about the informativeness of the modalities as a mask $M = [m^s, m^t, m^p, m^u]$, we propose an architecture to aggregate multi-modal embeddings to produce a single representation (i.e., embedding) for $r$ while exploiting the explicit knowledge. Here $m^{(.)} \geq 0$ is a weight to measure the informativeness of the corresponding modality in $r$, which is 0 for missing modalities in $r$. 

The GMU model initially transforms each multi-modal embedding using a modality-specific linear transformation matrix and followed by tanh activation as shown below:
\begin{align*}
    \tilde{z}^{(.)} &=\tanh (W^{(.)} \cdot z^{(.)})
    % \tilde{z}^t &=\tanh (W^t \cdot z^t)\\
    % \tilde{z}^p &=\tanh (W^p \cdot z^p)\\
    % \tilde{z}^u &=\tanh (W^u \cdot z^u)
\end{align*}
The aggregated representation is computed as the weighted average of the transformed multi-modal embedding. The weights assigned to each modality is computed as:
\begin{align}\label{eq:weighted_attention}
    w = softmax(M\odot W^w\cdot [z^s,z^t,z^p,z^u])
\end{align}
where $W^w \in \mathbb{R}^{4\times 4d}$ (assuming the original multi-modal embeddings are $d$-dimensional) and $softmax(x) = \frac{e^x}{\sum_{\forall j} e^{x_j}}$. The operators $\cdot$ and $\odot$ denote the dot product and the element-wise multiplication respectively. Eq.~\ref{eq:weighted_attention} can produce varying attention for different modalities based on their original multi-modal embeddings while exploiting the explicit knowledge input as a mask $m$.   

The final representation of $r$ is constructed as:
\begin{align}
    z = w \cdot [\tilde{z}^s,\tilde{z}^t,\tilde{z}^p,\tilde{z}^u]
\end{align}

For the rest of this manuscript, we denote the aforementioned whole operation inside a modified GMU model using the mapping function $GMU_\theta(.): [z^s,z^t,z^p,z^u, m]\rightarrow z$, which returns a low-dimensional vector to represent each new record using its multi-modalities. %(see Figure~\ref{fig:gmu_model}). 
Here, $\theta = \{W^s, W^t, W^p, W^u, W^w\}$ is the set of trainable parameters in the modified GMU model.

\subsection{Modality Agnostic Teacher-Student Architecture}\label{subsec:teacher-student}

\begin{figure}
    \centering
    \includegraphics[width=0.9\linewidth]{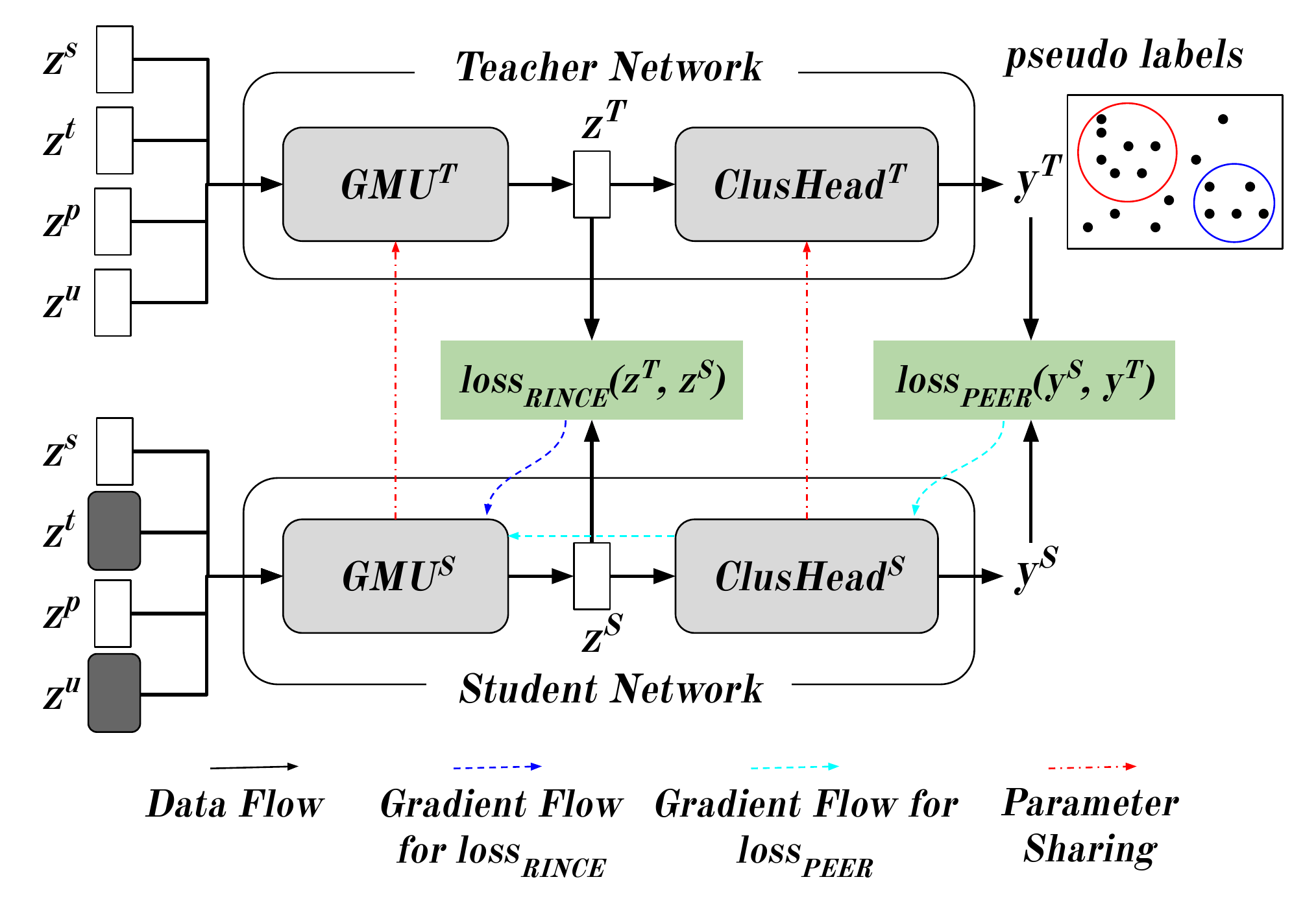}
    \caption{Overview of the teacher-student architecture -- both teacher and student share the same architecture consisting a Gated Multimodal Unit ($GMU$) to aggregate multimodal embeddings while emphasising informative modalities with the help of a noise-robust contrastive loss function ($loss_{RINCE}$) and $ClusHead$ to produce cluster labels from the multimodal news representations while alleviating the class-conditional label noise (CCN) using a CCN-robust loss function $loss_{PEER}$}.
    \label{fig:teacher_student_model}
    \vspace{-5mm}
\end{figure}

Here, we elaborate the proposed teacher-student architecture to predict veracity labels of news records while being agnostic to the available number of modalities for each news record. The proposed architecture is able to adopt the alignment of different modalities as a supervisory signal to preserve useful knowledge to identify fake news records without requiring a labelled dataset. As shown in Fig.~\ref{fig:teacher_student_model}, the teacher and the student networks in our technique share the same architecture -- each having the proposed GMU unit ($GMU^T$ and $GMU^S$) to generate the news representations $z^T$ and $z^S$ from the input multi-modal embeddings, and followed by a clustering head -- i.e., $ClusHead^T$ and $ClusHead^S$ to produce the soft cluster assignment of $r$ using the corresponding news representation $z^T$ and $z^S$ respectively. If we denote the parameters inside the cluster heads by $\xi$, the set of the trainable parameters in the teacher and student are $\{\theta^T, \xi^T\}$ and $\{\theta^S, \xi^S\}$ respectively. As can be seen in Fig.~\ref{fig:teacher_student_model}, our teacher network has access to all the modalities in $r$. However, the student network only has access to a subset of modalities. We can easily handle this difference during the training with the modified GMU unit in Section~\ref{subsec:multi-modal-agg}, by passing the mask with all ones for the teacher network and a mask with a set of zeros for the student network. For example, if we want to just mask the propagation speed-based modality (assuming it is the third modality), it can be achieved by passing $m$ as $[1, 1, 0, 1]$.

We only learn the parameters in our student network $\{\theta^S, \xi^S\}$ during training via back-propagation (please refer to the dotted arrows in Fig.~\ref{fig:teacher_student_model}), and update the parameters of the teacher networks $\{\theta^T, \xi^T\}$ as the exponential moving average of the parameters in the student~\cite{grill2020bootstrap}, which can be formally defined as follows:
\begin{align*}
    \theta^T \leftarrow \gamma  \theta^T + (1-\gamma)\theta^S\\
    \xi^T \leftarrow \gamma  \xi^T + (1-\gamma)\xi^S
\end{align*}
where we use a scheduler for $\gamma$ that linearly increases from $\gamma_0$ to $\gamma_n$ over the first $n$ updates and is kept constant for the remainder of the training. Such parameter sharing between the teacher and student architecture allows the student network to better mimic the operation of the teacher network with a partially available dataset.

\subsection{Noise-robust Misinformation Detection}\label{subsec:noise-robust-misinformation-detection}

In this section, we discuss two unsupervised and noise-robust objectives: $loss_{RINCE}$; $loss_{PEER}$, that we adopt to train the parameters in $(UMD)^2$. $loss_{RINCE}$ helps to imitate the performance of our teacher model using the student model with partial information. $loss_{PEER}$ helps to recover the knowledge in the news record embedding space $z$ to identify fake news in an unsupervised manner. 

\subsubsection{$loss_{RINCE}$} This loss function is motivated by the contrastive learning assumption -- the representations from positive pairs (e.g., the representations of the transformed views of the same image and the representations of an object from different modalities) should be similar compared to the negative pairs (e.g., the representations of the negative pairs). The conventional contrastive loss functions~\cite{li2021contrastive,chen2020simple} typically achieve this by maximizing similarity (e.g., dot product) between the representations of the positive pairs, and minimizing the similarity between the negative pairs. 

In our task, the representations of the teacher and the student models for the same news record are used to construct a set of positive pairs, and the teacher and the student representations from different records are used as negative pairs. Since some modalities are randomly masked for the input to the student network, some positive pairs in our task could be hard to make similar. For example, if the most informative modalities (e.g., source credibility) are masked in the input to the student networks, the representation of the student network could be quite different from the representation of the teacher network. Thus, applying the same weight for all positive pairs during learning could produce sub-optimal representations in the presence of such hard positive pairs. As a solution, the work in~\cite{chuang2022robust} proposed a noise-robust loss function, $loss_{RINCE}$, which could assign varying importance to different positive pairs according to their similarity. $loss_{RINCE}$ for a single training instance $\langle z_i^S, z_i^T, z^T_j\rangle$, consisting of the student and the teacher representations of the $i^{th}$ instance and the teacher representation of the $j^{th}$ $(i\neq j)$ instance, can be defined as follows:

\begin{equation}
    loss_{RINCE} = \frac{-e^{q\cdot s^+}}{q} + \frac{(\lambda\cdot (e^{q\cdot s^+} +e^{q\cdot s^-}))^q}{q}
\end{equation}
where $s^{+} = z_i^S \cdot z_i^T$ and $s^{-} = z_i^S \cdot z_j^T$. $q$ controls how the positive pairs should be weighted during the learning. When $q \rightarrow 0$, $loss_{RINCE}$ places more emphasise on hard positive pairs, and $q \rightarrow 1$ emphasises easy positive pairs more. To achieve the balance, we set $q=0.5$ following~\cite{chuang2022robust}.  
    
\subsubsection{$loss_{PEER}$} 
This loss term aims to train the clustering head
for predicting the cluster labels (e.g., veracity labels) of news records using their multi-modal representations -- e.g., $ClusHead^{(.)}: z^{(.)} \rightarrow \mathbb{R}^\kappa$, where $\kappa$ denotes the number of clusters. $loss_{PEER}$ is built on the notion that the veracity label of a new record $r$ from $ClusHead^T$ and $ClusHead^S$ should be consistent, despite some modalities of $r$ are not available for $ClusHead^S$. However, the learned multi-modal embeddings from $loss_{RINCE}$ are general, so they could include noisy information that is not informative for fake news detection. Thus, $loss_{PEER}$ adopts a pseudo labelling-based approach to alleviate noisy information iteratively while recovering useful knowledge to identify veracity labels of the records.

For each mini-batch of samples $R_B$, the soft-cluster assignment of the instances $P^T_B$ and $P^S_B$ are computed using both teacher and student networks. Then, the top $K$ confident instances $R^{'}_B$ are selected based on the assignment from the teacher network -- $R^{'}_B = \{i \in argtopk(max(P^T(i)))\}$, where $P^T(i)$ denotes the cluster assignment for the $i^{th}$ instance in $R_B$ from the teacher.

$loss_{PEER}$ is formulated to make consistent predictions from both $ClusHead^T$ and $ClusHead^S$ for the instances in $R^{'}_B$. Here, $loss_{PEER}$ considers the predictions from $ClusHead^T$ as labels  (or more accurate) since the teacher network has complete information about news records. However, some of the predictions from $ClusHead^T$ could be noisy. Thus, instead of minimizing the conventional cross-entropy loss between the predictions from each network, $loss_{PEER}$ is formulated as a peer loss~\cite{liu2020peer}, an extension of cross-entropy loss to make it noise-robust, which can be formally defined for a mini-batch as follows:
\begin{equation}
    loss_{PEER} = \frac{1}{|R^{'}_B|} \sum_{\forall j \in R^{'}_B}  l(P^S(j), Y(j)) - l(P^S(j^{'}), Y(j^{''})) 
\end{equation}
where $l(.)$ is the conventional cross-entropy loss and $Y(j) = argmax(P^T(j))$. $j^{'}$ and $j^{''}$ are two peer samples from $R^{'}_B$.

During learning, we linearly increase the size of the selected confident data pool -- i.e., $|R^{'}_B|$, by $5\%\times|R_B|$ steps starting from $10\%\times|R_B|$ until it covers the whole batch.  

\subsection{Inference}\label{subsec:inference}
For inference, $(UMD)^2$ provides two networks that makes consistent predictions: (1) the teacher network that predicts the labels of the news records with complete modality information; (2) the student network that makes predictions using a subset of the modalities with a mask consisting values for missing modalities as zero. %If a record has all the modalities available, the teacher network is adopted to make predictions. Otherwise, the student network can be adopted after setting the mask values for the missing modalities as zero. %with the corresponding mask -- e.g., if the propagation speed and user credibility modalities are missing, the student network can be used with $[1,1,0,0]$ mask. 

\section{Experimental Framework}\label{sec:main_experimental_framework}
\subsection{News Dataset Construction}\label{sec:news_dataset}
There are publicly available datasets for fake news detection -- e.g., PolitiFact~\cite{shu_fakenewsnet_2018} and GossipCop~\cite{shu_fakenewsnet_2018}. However, almost all these datasets are constructed by first collecting a set of fake news records using fact-checking services such as PolitiFact\footnote{\href{https://www.politifact.com/}{https://www.politifact.com/}} and Snopes\footnote{\href{https://www.snopes.com/}{https://www.snopes.com/}} and a set of real news records from reliable media outlets. Then, the tweets/retweets related to the selected news records are collected to generate social media content. We call this conventional approach as \textit{forward news dataset construction}. As shown in~\cite{zhou2021hidden}, this approach introduces several undesired artifacts/biases to the datasets. %, which hinders the generalizability of the models trained on such datasets. 

First, such a dataset may not reflect the actual statistics -- e.g., fake and real news proportions, of real-world news streams. For example, the proportion of fake and real news records in the PolitiFact dataset~\cite{shu_fakenewsnet_2018} is quite balanced despite the proportion of fake news records is typically small in a real-world news stream. Also, the fake and real news records of the same dataset may not fall in the same time-frame as they are collected from two different sources (Fig.~\ref{fig:distribution} (c)). Second, such a dataset is restricted to media-outlets that are either covered by fact-checking websites or known as reliable media sources. Hence, the coverage of different media outlets by this approach could be limited, which makes the scaling of such a dataset challenging. As a solution, we propose a novel approach to a large-scale dataset.
%to these limitations, this work proposes a novel approach to construct a large-scale news dataset for a given domain/event while addressing the following research questions:
% \begin{itemize}
%     \item RQ1: How can we maintain the empirical distribution of real and fake news records in a real-world news stream?
%     \item RQ2: How can we maximize the coverage of media outlets?
% \end{itemize}
We denote our approach \textit{backward news dataset construction} as it initially crawls tweets with news URLs related to the selected domain, then collects news articles mentioned in the tweets. The steps of \textit{backward news dataset construction} can be elaborated as follows:  
\begin{itemize}
    \item Step 1: Select a set of keywords for the selected domain/topic and a time-frame for the dataset.% collection.
    \item Step 2: Crawl all tweets (excluding retweets) posted during the selected time-frame that have at least one URL and one selected keyword.
    \item Step 3: Filter out tweets with non-news URLs using a news URL classifier (see %Section~\ref{subsec:news_url_classifier} 
    the supplementary material for more details).
    \item Step 4: Collect news articles and retweets/replies related to the remaining tweets.% to construct the dataset .
    \item Step 5: Filter out URLs that have $\leq C$ tweets/retweets (see Section~\ref{subsec:news_dataset} for more details about setting $C$).
\end{itemize}

\subsubsection{Very Large COVID-19 News Dataset}\label{subsec:news_dataset}

Following the steps in \textit{backward news dataset construction}, we construct \textsc{LUND-COVID}, a large-scale unlabelled news dataset related to the COVID-19 pandemic, to support future works on timely important COVID-19 misinformation detection. [Step 1] We select \textit{\{\#coronavirus, \#covid19, \#2019ncov, coronavirus, corona, covid19, pandemic, virus, wuflu\}} as the keywords and 2020/02/01 - 2020/07/15 as the time-frame. [Step 2] We collected around 14 million unique URLs posted via tweets during this time-frame that have at least one selected keyword. [Step 3] After filtering out non-news URLs, 7,590,329 unique news URLs belonging to 27,907 media-outlets are identified. [Step 4] After collecting both tweets and retweets of these URLs, the dataset ends up with 30,123,010 tweets. [Step 5] To remove URLs with fewer tweets, we set the threshold value $C$ to 10 as it yields similar average tweets per URL in our dataset to the CoAID dataset~\cite{cui2020coaid}, a widely used COVID-19 related news dataset collected by \textit{forward news dataset construction}. At the end of this final filtering stage, our dataset consists of 419,351 URLs and 17,802,652 tweets including retweets. 

\begin{figure}[t]
    \begin{center}
        \resizebox{\linewidth}{!}{\input{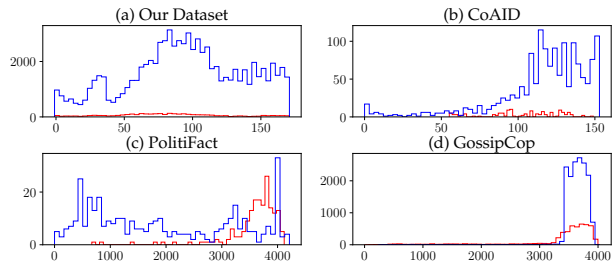}}
    \end{center}
    \vspace{-5mm}
    \caption{The distribution of fake (red plots) and real news (blue plots) in different datasets -- x and y axes represent the timespan of each dataset in days and the number of news records from each class respectively. The labels for our dataset is produced using the weak labelling approach proposed in~\cite{zhou2020recovery}.}
    \label{fig:distribution}
    \vspace{-1mm}
\end{figure}

\begin{table}[t]
    \centering
    \scriptsize
    \begin{tabular}{|c|c|c|c|c|}
    \hline 
         Dataset & LUND-COVID & CoAID & PolitiFact & GossipCop  \\\hline\hline
         \# domains & 19,590 & 127 & 354 & 1757 \\\hline
         \# articles per domain & 10.8 & 16.8 & 2.6 & 9.7 \\\hline
         \# domains per day & 115.2 & 0.8& 0.1&0.4 \\\hline
    \end{tabular}
    \vspace{3mm}
    \caption{The statistics related to media-outlet coverage of different datasets.} % from three perspectives: (1) the total number of media-outlets; (2) the number of articles for a single media-outlet; and (3) the number of unique media-outlets per day in the timespan.}
    \label{fig:domain_coverage}
    \vspace{-8mm}
\end{table}

In Figure~\ref{fig:distribution}, we evaluate the distribution of fake and real news in our \textsc{LUND-COVID} dataset collected using \textit{backward news dataset construction} with three publicly available datasets: (1) CoAID~\cite{cui2020coaid}; (2) PolitiFact~\cite{shu_fakenewsnet_2018}; and (3) GossipCop~\cite{shu_fakenewsnet_2018}, constructed from \textit{forward news dataset construction}. Since our dataset does not have labels, we adopt the approach proposed in~\cite{zhou2020recovery} for this analysis to produce labels for our datasets. It can be clearly seen that our approach consistently crawls real and fake news records across the whole dataset timespan, which ultimately helps to maintain the empirical distribution of real and fake news records in the real-world. However, the conventional \textit{forward news dataset construction} cannot guarantee this property -- e.g., Fig.~\ref{fig:distribution} (c) which has a cluster of fake news towards the end of the timespan. In our dataset, we only observe $\sim 1\%$ of the records as fake news. This figure nearly aligns with the fake news proportions reported in previous analysis, which further verifies the potential of \textit{backward news dataset construction} to preserve the empirical statistics of fake and real news. However, the same figure is much higher for almost all the other datasets (see Fig.~\ref{fig:distribution} (c,d)).  

% \begin{figure}
%     \begin{center}
%         \resizebox{0.75\linewidth}{!}{\input{histogram_2.pgf}}
%     \end{center}
%     \vspace{-5mm}
%     \caption{The media-outlet coverage of different datasets from three perspectives: (1) the total number of media-outlets; (2) the number of articles for a single media-outlet; and (3) the number of unique media-outlets per day in the timespan. Due to the significant differences in the range of each metric, each metric is plotted in the natural logarithmic scale.}
%     \label{fig:domain_coverage}
%     \vspace{-3mm}
% \end{figure}

%\subsubsection{Media outlet Coverage}
Table~\ref{fig:domain_coverage} analyses the media outlet coverage of our dataset compared to the existing datasets. As can be seen, the total number of different media outlets covered by our dataset is much higher than the existing datasets. This could be due to the larger number of records in our dataset. Thus, we report the number of records for each media-outlet in Table~\ref{fig:domain_coverage}, which shows a higher number of records compared to most existing datasets such as PolitiFact and GossipCop. %These results verify the potential of \textit{backward news dataset construction} to maximize the media outlet coverage by capturing the media outlets that are not typically fact-checked by fact-checking websites. 

The aforementioned results using \textsc{LUND-COVID} verify that the proposed \textit{backward news dataset construction} technique can produce a news dataset that reflects the statistics of a real-world news stream. We adopt \textsc{LUND-COVID} to train our weak source embeddings modules and $(UMD)^2$. Since we do not have clean veracity labels for the articles in \textsc{LUND-COVID}, the proposed unsupervised framework is evaluated using three publicly available labelled datasets, which are relatively small compared to our dataset (see Table~\ref{tab:statistics}). Since our training dataset consists of COVID-19 related news records, we perform an in-domain evaluation using CoAID~\cite{cui2020coaid}, a labelled dataset related to COVID-19 misinformation. To evaluate the generalizability of our approach for the unseen domains during training, we evaluate against two out-of-domain datasets: (1) PolitFact~\cite{shu_fakenewsnet_2018}, related to Politics; and (2) GossipCop~\cite{shu_fakenewsnet_2018} related to Entertainment.  % Table~\ref{tab:statistics} compares the statistics of the selected evaluation datasets against our dataset.

%Our next problem is to produce the veracity labels for the news records in an unsupervised manner. %To address this, we propose a self-supervised objective function that relies on various weak sources of news records. We first formally define this problem in Section~\ref{sec:problem_statement}, which will be discussed in detail in Section~\ref{sec:embedding} and Section~\ref{sec:inference}.
\begin{table*}[t]
    \scriptsize
    \centering
    %\caption{Results for fake news detection of different methods, which are classified under three categories: (1) text content-based approaches (T); (2) social context-based approaches (S); and (3) multi-modal approaches (M).}%; and (3) suitability for early fake news detection (i.e., ability to yield better performance with initial propagation networks) (E)}
    \begin{tabular}{|p{3.8cm}|p{0.15cm}|p{0.15cm}|p{0.15cm}|p{0.15cm}|p{0.5cm}|p{0.5cm}|p{0.5cm}|p{0.5cm}|p{0.5cm}|p{0.5cm}|p{0.5cm}|p{0.5cm}|p{0.5cm}|p{0.5cm}|p{0.5cm}|p{0.5cm}|}
        \hline
         Method & \multicolumn{4}{c|}{Modalities}&\multicolumn{4}{c|}{CoAID}& \multicolumn{4}{c|}{Politifact} &\multicolumn{4}{c|}{Gossipcop}\\ 
         \cline{2-17}
         &S&T&P&U& Acc & Prec & Rec & F1& Acc & Prec & Rec & F1 &Acc & Prec & Rec & F1\\
         \hline
         Source credibility-based $UMD$ &\checkmark&&&&0.852&0.595&0.815&0.618&0.724&0.715&0.707&0.710&0.662&0.541&0.627&0.568\\
         News content-based $UMD$ &&\checkmark&&&0.518&0.504&0.523&0.515&0.717&0.709&0.703&0.704&0.718&0.704&0.708&0.705\\
         Propagation speed-based $UMD$ &&&\checkmark&&0.701&0.568&0.842&0.579&0.807&0.709&0.678&0.691&0.701&0.696&0.690&0.692\\
         User credibility-based $UMD$ &&&&\checkmark&\bf 0.930&0.490&0.495&0.491&0.701&0.682&0.686&0.683&0.722&0.719&0.717&0.717\\
         \hline
         Majority Voting &\checkmark&\checkmark&\checkmark&\checkmark&0.781&0.573&0.817&0.581&0.718&0.705&0.702&0.703&0.728&0.708&0.711&0.708\\
         UFNDA &&\checkmark&&&0.574&0.520&0.468&0.475&0.685&0.667&0.659&0.670&0.692&0.687&0.662&0.673\\
         TruthFinder &&\checkmark&&\checkmark&0.769&0.476&0.501&0.488&0.581&0.572&0.576&0.573&0.668&0.669&0.672&0.669\\
         UFD &&\checkmark&&\checkmark&-&-&-&-&0.697 & 0.652& 0.641& 0.647&0.662 &0.687&0.654&0.667\\ 
         Recovery &\checkmark& \checkmark&&&0.831&0.493&0.708&0.511&0.711&0.693&0.685&0.686&0.660&0.330&0.500&0.398\\
         GTUT &\checkmark&\checkmark&&\checkmark&0.882&0.571&0.792&0.650&0.776&0.782&\bf 0.758&\bf 0.767&0.771& 0.770&0.731&0.744\\
        %  Cooperative Learning &\checkmark&\checkmark&\checkmark&\checkmark& 0.892&0.599&0.801&0.679&0.754&0.771&0.729&0.738&0.749&0.740&0.748&0.743\\
         \hline
         \hline
         $(UMD)^2$&\checkmark&\checkmark&\checkmark&\checkmark& 0.917&\bf 0.668&\bf 0.849&\bf 0.708&\bf 0.802&\bf 0.795&0.748&0.761&\bf 0.792&\bf0.779&\bf 0.788&\bf 0.783\\
         \hline
         \multicolumn{5}{|l|}{\textbf{Ablation Study}} &&&&&&&&&&&&\\
         \multicolumn{5}{|l|}{{$(UMD)^2$ (-) \textit{$loss_{RINCE}$}}}&0.897&0.608&0.822&0.694&0.788&0.772&0.742&0.749&0.786&0.768&0.773&0.770\\
         \multicolumn{5}{|l|}{{$(UMD)^2$ (-) \textit{$loss_{PEER}$}}}&0.905&0.648&0.798&0.691&0.792&0.779&0.747&0.754&0.783&0.772&0.765&0.768\\
         \hline
    \end{tabular}
    \vspace{3mm}
    \caption{Results for unsupervised fake news detection of different methods, which are classified under the modalities they use: (1) source credibility-based (S); news content-based (T); (2) propagation speed-based (P); and (3) user credibility-based (U).}
    \label{tab:results0}
    \vspace{-5mm}
\end{table*}

\subsection{Baselines}
We adopt a set of unsupervised uni-modal approaches (denoted as $UMD$): (1) source credibility-based $UMD$; (2) news content-based $UMD$; (3) propagation speed-based $UMD$; and (4) user credibility-based $UMD$. For each baseline, the pre-trained embeddings from the weak source are clustered using agglomerative clustering\footnote{We evaluated the performance of $UMD$ using other clustering algorithms such as K-means and spectral clustering, and observed that agglomerative clustering consistently yielded better performance compared to the others. Thus, the agglomerative clustering algorithm was adopted for the reported results for $UMD$} to produce the labels.  

In addition, we adopt six widely-known baselines for unsupervised fake news detection: Majority Voting, UFNDA~\cite{li2021unsupervised}, TruthFinder~\cite{yin2008truth}, Recovery~\cite{zhou2020recovery}, UFD~\cite{yang2019unsupervised} and GTUT~\cite{gangireddy2020unsupervised}. See our supplementary material for more details about the baselines. Also, we compare $(UMD)^2$ with two weaker variants for an ablation study: 
\begin{itemize}
    \item $(UMD)^2\; (-)\; loss_{RINCE}$ replaces the $loss_{RINCE}$ loss term using the $loss_{SIMCLR}$ loss function~\cite{li2021contrastive}.
    \item $(UMD)^2\; (-)\; loss_{PEER}$ replaces the $loss_{PEER}$ loss term using the cross-entropy loss function.
\end{itemize}

\subsection{Evaluation Metrics}
$(UMD)^2$ returns cluster assignment for the news records in the datasets. Following the conventional experimental setup for clustering, we adopt the Hungarian algorithm~\cite{bruff2005assignment} to map clusters to the labels. This evaluation strategy is consistent for all the baselines. After mapping each cluster to a best label, we adopt Accuracy, Precision, Recall, and F1-score as the evaluation metrics.

\section{Results}\label{sec:results}

This section discusses the performance of the proposed approach for fake news detection under three settings: (1) unsupervised fake news detection -- this task assumes that all the modalities are available during training and testing, but not the labels; (2) unsupervised fake news early detection -- this task exploits all the modalities during training, but assumes that only source credibility and textual content are available during testing; and (3) few-shot fake news detection -- this task requires a few clean veracity labels  (e.g., 5-shot learning requires 5 clean labels) during training. We elaborate these settings further in the subsequent sections along with the results with each setting.

\begin{table*}[t]
    \scriptsize
    \centering
    %\caption{Results for fake news detection of different methods, which are classified under three categories: (1) text content-based approaches (T); (2) social context-based approaches (S); and (3) multi-modal approaches (M).}%; and (3) suitability for early fake news detection (i.e., ability to yield better performance with initial propagation networks) (E)}
    \begin{tabular}{|p{3.8cm}|p{0.15cm}|p{0.15cm}|p{0.15cm}|p{0.15cm}|p{0.5cm}|p{0.5cm}|p{0.5cm}|p{0.5cm}|p{0.5cm}|p{0.5cm}|p{0.5cm}|p{0.5cm}|p{0.5cm}|p{0.5cm}|p{0.5cm}|p{0.5cm}|}
        \hline
         Method & \multicolumn{4}{c|}{Modalities}&\multicolumn{4}{c|}{CoAID}& \multicolumn{4}{c|}{Politifact} &\multicolumn{4}{c|}{Gossipcop}\\ 
         \cline{2-17}
         &S&T&P&U& Acc & Prec & Rec & F1& Acc & Prec & Rec & F1 &Acc & Prec & Rec & F1\\
         \hline
         Source credibility-based $UMD$ &\checkmark&&&&0.852&0.595&0.815&0.618&0.724&0.715&0.707&0.710&0.662&0.541&0.627&0.568\\
         News content-based $UMD$ &&\checkmark&&&0.518&0.504&0.523&0.515&0.717&0.709&0.703&0.704&0.718&0.704&0.708&0.705\\
         \hline
         Majority Voting &\checkmark&\checkmark&&&0.801&0.581&0.802&0.597&0.732&0.718&0.709&0.714&0.706&0.685&0.698&0.692\\
         UFNDA &&\checkmark&&&0.574&0.520&0.468&0.475&0.685&0.667&0.659&0.670&0.692&0.687&0.662&0.673\\
         Recovery &\checkmark& \checkmark&&&0.831&0.493&0.708&0.511&0.711&0.693&0.685&0.686&0.660&0.330&0.500&0.398\\
         % Cooperative Learning &\checkmark&\checkmark&&& &&&&&&&&&&&\\
         \hline
         \hline
         $(UMD)^2$&\checkmark&\checkmark&&& \bf 0.892&\bf 0.631&\bf 0.824&\bf 0.692&\bf 0.798&\bf 0.787&\bf0.732&\bf 0.752&\bf 0.743&\bf0.741&\bf 0.754&\bf 0.746\\
         \hline
    \end{tabular}
    \vspace{3mm}
    \caption{Results for unsupervised fake news early detection. S, T, U and P denote source credibility-based, news content-based, propagation-based and user credibility-based modalities, respectively.}
    \label{tab:results1}
    \vspace{-7mm}
\end{table*}

\subsection{Results for Unsupervised Fake News Detection}
In this task, we assume that the information about all the modalities is present for the test instances. Thus, the teacher network in $(UMD)^2$ is used for inference.  

As shown in Table~\ref{tab:results0}, the proposed approach yields substantially better results for both in-domain and out-of-domain datasets. Out of the baselines, GTUT yields the best results for most figures. GTUT requires the presence of the instances in the evaluation datasets during training without their labels. However, most other baselines including our model do not require that. Thus, GTUT has a slight advantage over other models in Table~\ref{tab:results0}, which could be a reason for its relatively strong performance. However, our approach still outperforms GTUT for most datasets. For CoAID, $(UMD)^2$ outperforms GTUT by 8.9\% in F1-score. In the out-of-domain evaluation, $(UMD)^2$ yields competitive performance with GTUT for PolitiFact and achieves around 5\% F1-score improvements for GossipCop. These results verify the generalizability of our approach to unseen domains during training as $(UMD)^2$ is trained only using a COVID19-related dataset. 

$(UMD)^2$ exploits 4 different modalities in this work. However, it can be easily extended for other modalities. All the baselines in Table~\ref{tab:results0} except Majority Voting cannot exploit all the available modalities. Table~\ref{tab:results0} shows that Majority Voting is unable to outperform the strongest uni-modal baseline in most datasets. This could be due to its inability to identify informative weak sources and to weigh them accordingly. In contrast, our model can emphasise informative modalities using its attention mechanism. 

Furthermore, we analyse the contribution of different loss terms in $(UMD)^2$. It can be seen that both $Loss_{RINCE}$ and $Loss_{PEER}$ terms positively contribute towards the final performance. As can be seen in Section~\ref{sec:embedding}, the selected modalities and their pre-trained embedding spaces are not perfect for fake news detection. Thus, they carry noisy information. $Loss_{RINCE}$ and $Loss_{PEER}$ terms are robust to such noise, which could be the reason for their positive contribution compared to their standard counterparts $Loss_{SIMCLR}$ and $Loss_{Cross\;Entropy}$ respectively. 

\subsection{Results for Unsupervised Fake News Early Detection}
This task assumes that propagation speed and user-credibility modalities are unavailable for inference. Thus, the teacher network in $(UMD)^2$ is trained with full information and the student network is trained without propagation speed and user-credibility modalities. As explained in Section~\ref{subsec:inference}, the test labels are predicted using the student network. Most existing baselines (e.g., UFD and GTUT) are unable to work under this setting as they are built on the modalities (e.g., user credibility) that are unavailable for fake news early detection. As shown in Table~\ref{tab:results1}, our model outperforms the workable baselines under this setting by as much as 12\% in F1-score. Thus, $(UMD)^2$ effectively addresses the challenge of having missing modalities via its teacher-student architecture. See our supplementary material for more experiments with missing modalities.

\subsection{Results for Few-shot Fake News Detection}
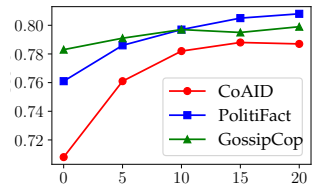
\begin{figure}[t]
    \begin{center}
    \resizebox{0.5\linewidth}{!}{\input{histogram_3.pgf}}
    \end{center}
    \vspace{-5mm}
    \caption{F1-score of the proposed model $(UMD)^2$ for k-shot fake news detection task under different k values.}
    \label{fig:few-shot}
    \vspace{-5mm}
\end{figure}

Under this setting, we assume that the clean labels for a few selected instances $k$ (e.g., 5 or 5-shot learning) can be produced via a human annotator. Thus, we set the number of clusters in $ClusHead^T$ and $ClusHead^S$ for $k$. Then the label for an instance from each cluster is taken via the annotator in the loop and propagates the same label for other instances in the same cluster. Since some of the fake news and real news from different domains have distinct characteristics, mapping them into multiple small clusters instead of mapping to two big clusters could help to preserve such cross-domain differences. Figure~\ref{fig:few-shot} verifies this intuition by showing improved performance of $(UMD)^2$ for larger $k$ values. However, the performance of $(UMD)^2$ for the $k$-shot fake news detection task reaches a plateau when $k$ increases -- e.g., $k>10$ values yield consistent performance for CoAID. This observation verifies that the performance of $(UMD)^2$ can be further improved with a minimal effort from human fact-checkers. 

\section{Related Work}\label{sec:related_work}
\subsection{Multi-modal Fake News Detection}
Fake news detection methods rely on different modalities (text, source, social context) of news records to determine their veracity. Text content-based approaches~\cite{yang_hierarchical_2016,perez-rosas_automatic_2018,pennebaker_development_2015} explore word usage and linguistic styles in the headline and body of news records to identify fake news. Some works explore the credibility of the source of the articles~\cite{nakov2020can,zhou2020recovery}. Also, some works~\cite{silva2021propagation2vec,shu_hierarchical_2019,silva-embedding-2020,monti_fake_2019,wu_tracing_2018,ma_detect_2017} exploit the features from the social context (e.g., the features of the propagation pattern and the characteristics of the users engaging with the news records) of news records to identify their veracity with the help of machine learning techniques such as Propagation Tree Kernels~\cite{ma_detect_2017}, Recurrent Neural Networks~\cite{silva2021propagation2vec,wu_tracing_2018}, and Graph Neural Networks~\cite{monti_fake_2019}. These works verify the existence of considerable differences of these various modalities for real and fake news. Motivated by this finding, several works attempted to jointly exploit several of such modalities~\cite{silva2021embracing,shu_leveraging_2020,wang_eann_2018,shu_defend_2019} together for improved fake news detection. Our approach belongs to this category. These multi-modal approaches can outperform uni-modal approaches due to their ability to exploit informative knowledge from various sources~\cite{silva2021embracing,shu_leveraging_2020}. However, almost all the existing multi-modal approaches require clean labels at least for a few training instances~\cite{shu_leveraging_2020}. In contrast, our model does not require clean labels. Instead, our approach exploits the alignment of various multi-modal signals as a supervisory signal to train the model, which is not well-studied in the previous works on fake news detection. 

\subsection{Unsupervised Fake News Detection}
There have been a few previous attempts~\cite{yin2008truth,yang2019unsupervised,li2021unsupervised} to detect fake news in an unsupervised manner. TruthFinder~\cite{yin2008truth} is one of the earliest works in this line, which considers the relationship between multiple websites and the content presented on these websites to determine the truthfulness of a story. TruthFinder can be fooled due to coordinated news sharing behaviour in today's media outlets (see Section~\ref{subsec:source_credibility} for more details). The study in~\cite{li2021unsupervised} learns an autoencoder using real news records such that it returns a higher reconstruction error for fake news. This model represents each news article using its textual and image attributes. With the emergence of social media, it has been found that today people mostly use social media to seek news, thus the propagation pattern via social media has been identified as an informative source to identify fake news. The works in~\cite{yin2008truth,li2021unsupervised} are unable to exploit such complex, yet informative, modalities. As a solution, the work in~\cite{yang2019unsupervised} recently proposed UFD, a graphical framework based on Bayesian principles to represent the dependencies among the truths of news, the users’ opinions, and the users’ credibility, which is then used to infer the veracity of the articles using a modified Gibbs sampling approach. The scalability and online learning capability are quite limited in this approach, despite of its good and explainable performance for fake news detection. GTUT~\cite{gangireddy2020unsupervised} proposed another graph-based approach, which exploits the inter-user behaviour in news propagation via a label propagation mechanism to identify fake news. Almost all these approaches focus on a particular modality in news records (i.e., uni-modal) and also it is difficult to extend them to other modalities. In contrast, our approach studies how to exploit multi-modalities in an unsupervised manner for improved misinformation detection. Our approach is designed in a manner that it can be easily extended for other modalities. In this work, the proposed unsupervised fake news detection framework is quantitatively compared against 
the aforementioned baselines~\cite{yin2008truth,yang2019unsupervised,li2021unsupervised,gangireddy2020unsupervised}. Our results also show that our approach considerably outperforms the existing baselines, which further verifies the importance of the contributions of the proposed framework.% in unsupervised fake news detection.  

\section{Conclusion}\label{sec:conclusion}
To effectively exploit the multi-modal knowledge available in news datasets in an unsupervised manner, we propose a novel unsupervised fake news detection framework, which first encodes the knowledge available in various modalities across the lifespan of a news record -- source, text, propagation, and users, as low-dimensional vectors in a domain-agnostic manner, and then passes them through a noise-robust teacher-student architecture to identify the veracity labels by exploiting the alignment of the different modalities as a supervisory signal. Also, this work proposes a novel new dataset collection technique, \textit{backward news dataset construction}, using which we can generate large-scale multi-modal news datasets while minimizing the latent biases in the standard news datasets. Following \textit{backward news dataset construction}, we generate \textsc{LUND-COVID}, a large-scale multi-modal news dataset on COVID-19 consisting of more than 400,000 news articles. The proposed unsupervised framework is trained using \textsc{LUND-COVID} and tested using one in-domain and two out-of-domain datasets. The results show that the proposed framework consistently outperforms existing baselines by as much as 12\% in F1-score, while showing better generalizability for unseen domains during training. Our results on fake news early detection task show that the proposed approach is robust against the missing modalities during testing. Also, the performance of our approach can be further boosted with the presence of a few clean labels.

For future work, we intend to extend our model for other modalities such as images and user reactions. Since the evaluation in this work is offline, the performance of our model in an online environment is another interesting direction to explore. This setting introduces new challenges such as speed and memory optimization for fitting into online frameworks.

\bibliographystyle{IEEEtran}
\bibliography{references}

\begin{IEEEbiography}[{\includegraphics[width=1in,height=1.25in,clip,keepaspectratio]{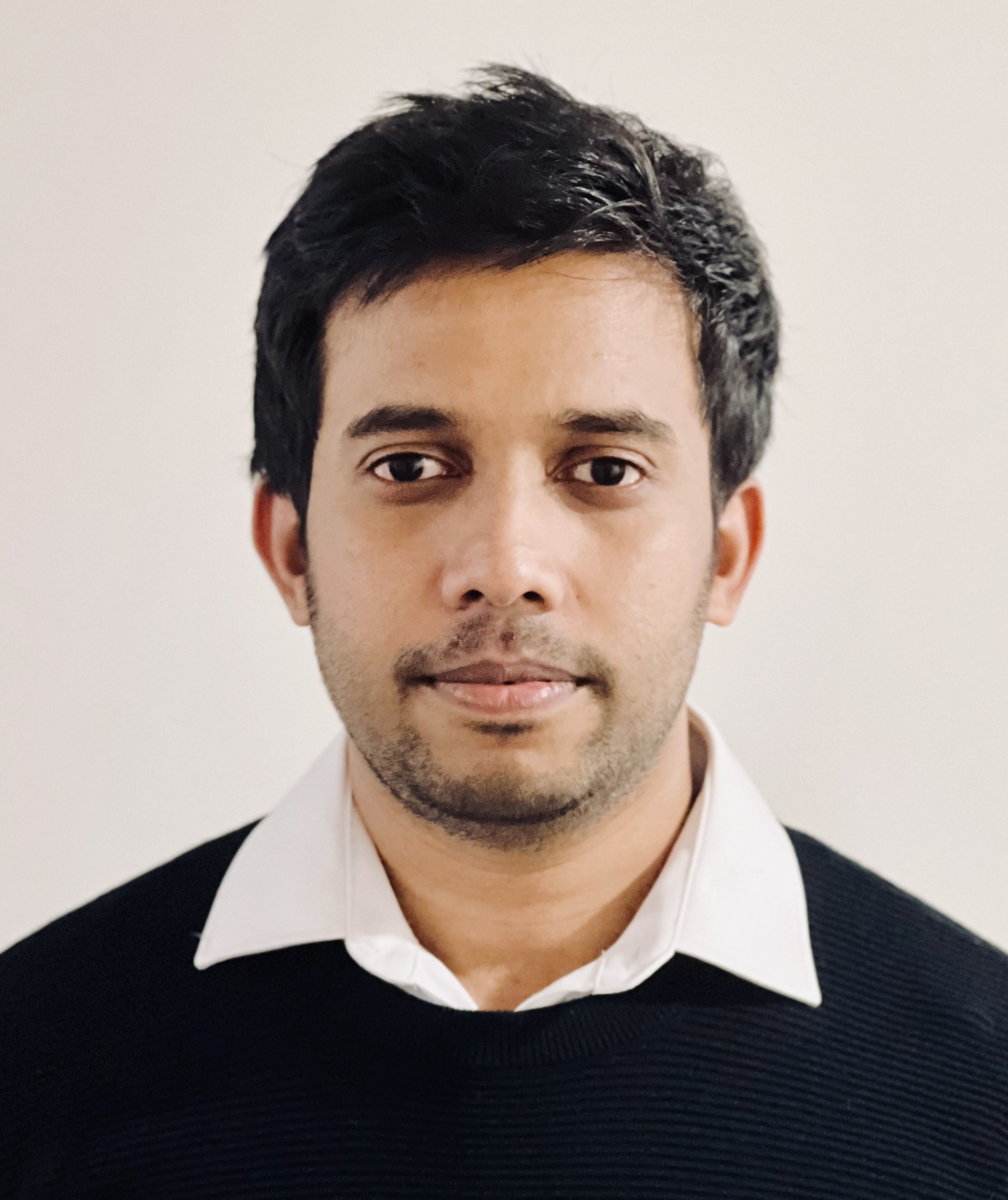}}]{Amila Silva} is a Ph.D. student at the School of Computing and Information Systems, University of Melbourne, Australia. He received his B.Sc. in electronics and
telecommunications engineering from the University of Moratuwa, Sri Lanka, in 2017. His research interests lie in multimodal machine learning, representation learning and fake news detection. He has published over 10 papers in conferences and journals including AAAI, ECML-PKDD, CIKM, PAKDD etc.
\end{IEEEbiography}

\begin{IEEEbiography}[{\includegraphics[width=1in,height=1.25in,clip,keepaspectratio]{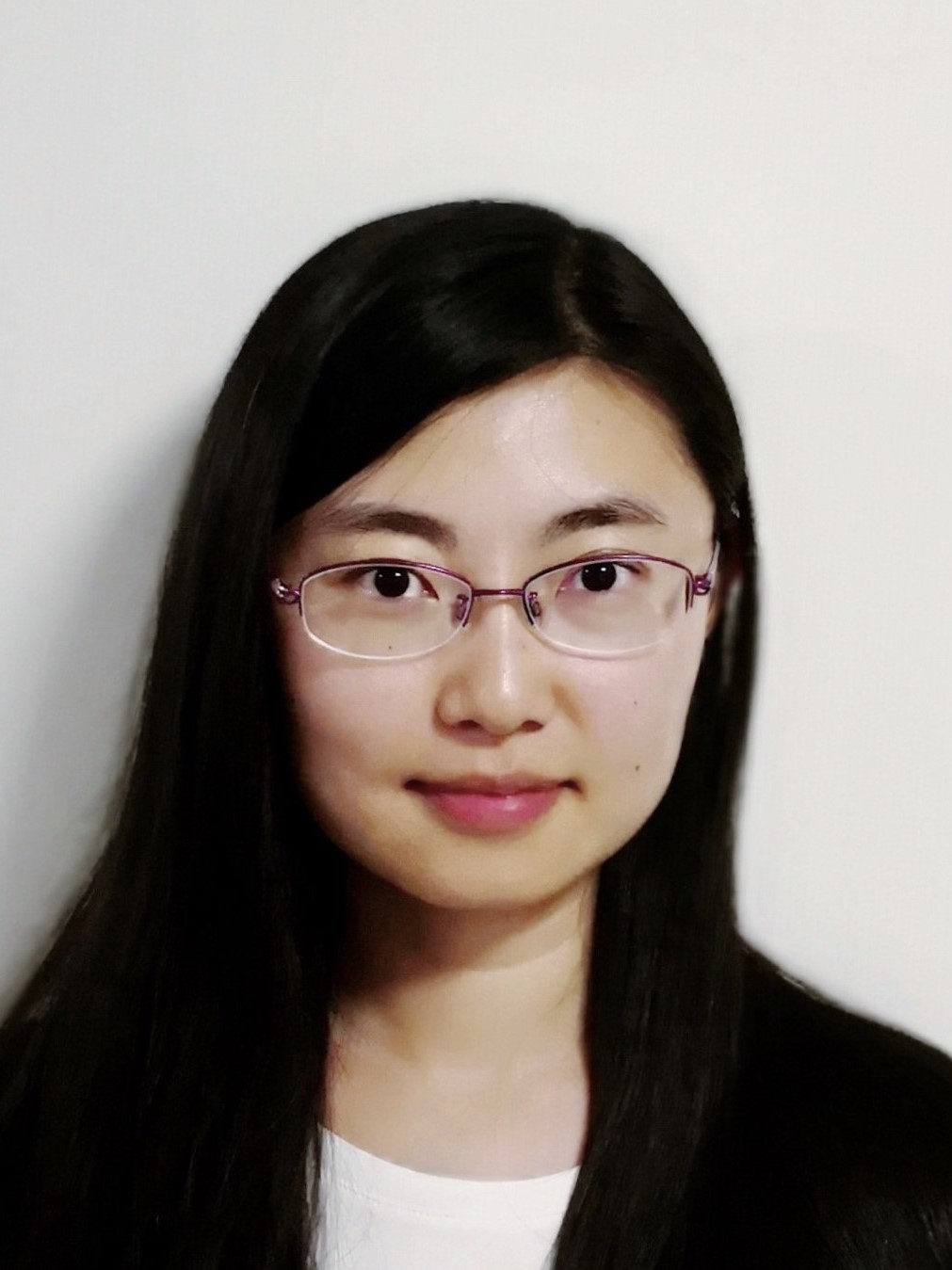}}]{Ling Luo}
is a Lecturer at the School of Computing and Information Systems, University of Melbourne, Australia. Ling received PhD in data mining and machine learning from the University of Sydney in 2017. Ling’s research interests include temporal modeling, user behavior analytics and stochastic processes.
\end{IEEEbiography}

\begin{IEEEbiography}[{\includegraphics[width=1in,height=1.25in,clip,keepaspectratio]{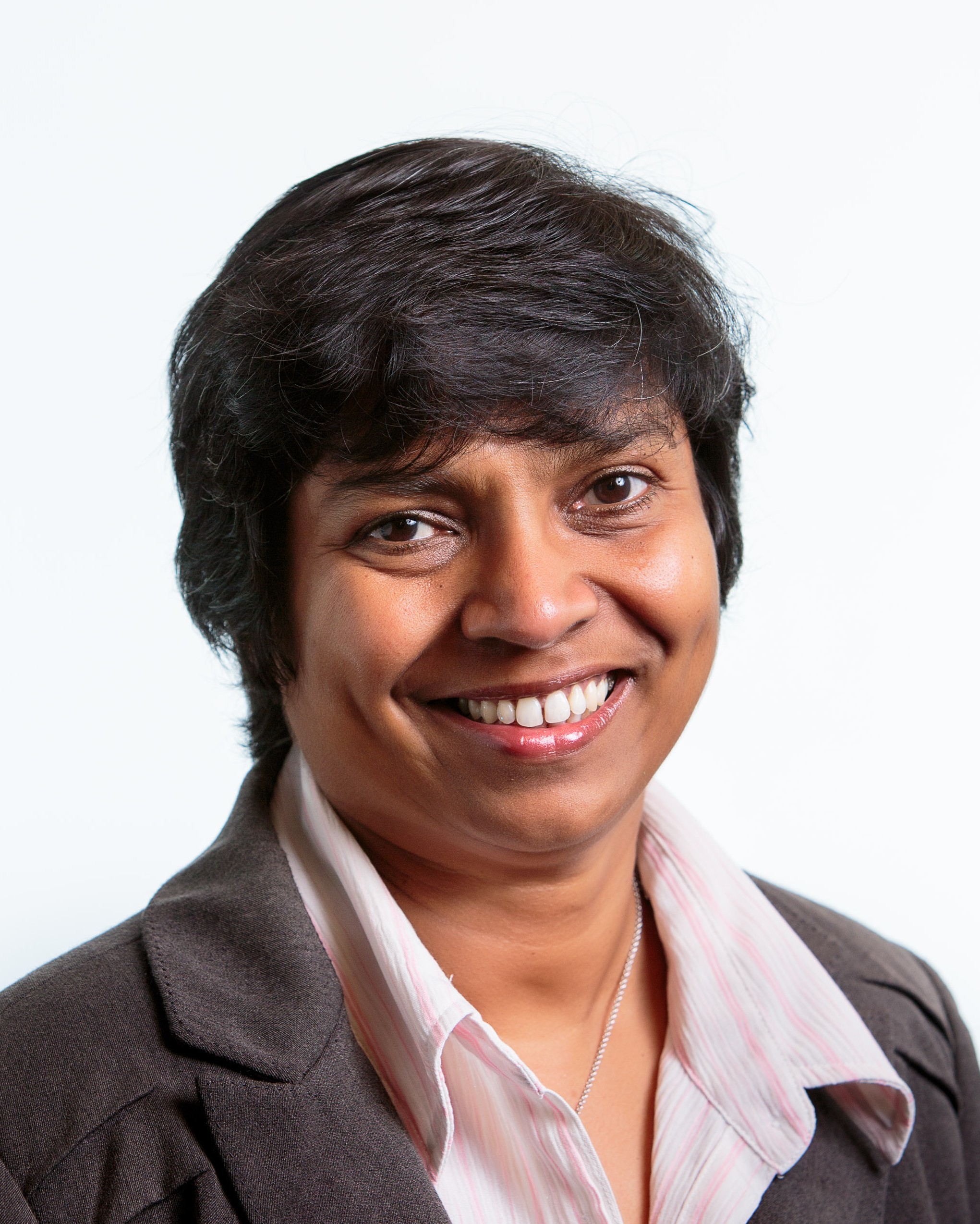}}]{Shanika Karunasekera} received the B.Sc. degree in electronics and telecommunications engineering from the University of Moratuwa, Sri Lanka, in 1990 and the Ph.D. degree in electrical engineering from the University of Cambridge, Cambridge, U.K., in 1995. From 1995 to 2002, she was a Software Engineer and a Distinguished Member of Technical Staﬀ with Lucent Technologies, Bell Labs Innovations, USA. In January 2003, she joined the University of Melbourne,  Victoria, Australia, and currently she is a Professor in the Department of Computing and Information Systems. Her current research interests include distributed system engineering, machine learning, data-mining and social media analytics.
\end{IEEEbiography}

\begin{IEEEbiography}[{\includegraphics[width=1in,height=1.25in,clip,keepaspectratio]{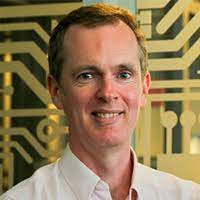}}]{Christopher Leckie} received a B.Sc. degree, a B.E. degree in electrical and computer systems engineering (with first class honours), and a Ph.D.  degree in computer science  from Monash University, Australia, in 1985, 1987, and 1992,  respectively.  He is currently a Professor with the School of Computing and Information Systems at the University of Melbourne. His research interests include scalable data mining and security analytics.
\end{IEEEbiography}



\section*{Supplementary material}

# Supplementary Material for Unsupervised Domain-agnostic Fake News Detection using Multi-modal Weak Signals

Amila Silva, Ling Luo, Shanika Karunasekera, and Christopher Leckie

**Abstract**—This is the supplementary material for the manuscript titled "Supplementary Material for "Unsupervised Domain-agnostic Fake News Detection using Multi-modal Weak Signals".

**Index Terms**—fake news detection, unsupervised learning, weak signals

✦

## 1 Representing Affective Information in Text

To represent the affective information of a news record that are invariant across domains, our work adopts the following steps:

- For a given news article $r$, we combine its title and body text as its textual content.
- The textual content is first represented as a bag of words (BoW), where the feature value of each word is its normalized term frequency using the articles' length.
- The following 111 high-level features are extracted using the BoW representation and the original text to represent the article's affective information $t^r$:
  - *Sentiment:* We extract the sentiment from the text using a fine-tuned version of roBERTa [1], the widely-known pretrained lanugage model, for sentiment classification (1 Features).
  - *Emotions:* To detect emotional differences among articles, we adopt the NRC emotions lexicon [2] that contains around 14K words labeled using the eight Plutchik's emotions (8 Features).
  - *Psycholinguistic:* The categories from the LIWC dictionary [3] are used to represent the psycholinguistic differences of the articles (90 Features).
  - *Readability:* To measure the readability of the articles, SMOG readability measure [4] is adopted (1 Feature).
  - *Morality:* We consider cue words from the Moral Foundations Dictionary [5] where words are assigned to one (or more) of the following categories: care, harm, fairness, unfairness (cheating), loyalty, betrayal, authority, subversion, sanctity and degradation (10 Features).
  - *Hyperbolic:* We adopt the dictionary proposed in [6] that includes around 350 eye-catching words (e.g., breathtakingly, terrifying, soul-stirring etc.) to identify the proportion of such words in the article (1 Feature).

Our pretraining embedding function for textual content adopts the aforementioned feature vector of a news record as the input to the embedding function.

## 2 Social Engagement-based Graph Construction

In our proposed approach for embedding user credibility, the knowledge in the social engagements ($u^r$) of a news record $r$ is initially converted into a network structure (i.e., *engagement-based graph* and $G^u$), denoted using a tuple – $G^u = (V^u, A^u, X^u)$, where:

- $V^u$ represents the set of users who engaged with $r$ via tweeting or retweeting the news article $r$. There is a special node $v^u_\star$ in $V^u$ to represent the news article itself. $|V^u|$ denotes the total number of nodes in $G^u$.
- $A^u \in \mathbb{R}^{|V^u| \times |V^u|}$ represents the adjacency matrix of $G^u$ – $A^u(i,j) = 1$ if there is an edge between $v^u_i$ and $v^u_j$; 0 otherwise. There is an edge from node $v^u_i$ to node $v^u_j$ if (1) the user of tweet $v^u_i$ mentions the user of tweet $v^u_j$; or (2) the tweet posted by $v^u_i$ is public and the tweet posted by user $v^u_j$ is posted within a certain period. For user tweets that are not satisfied, at least one of the previous two conditions are connected to news record node $v^u_\star$ to form a new cascade in $G^u$.
- $X^u \in \mathbb{R}^{|V^u| \times d}$ represents the features of the nodes. Each user node $v^e_i$ is characterized using d=10 different features in the corresponding user profiles/tweets that have been shown to be useful to detect user credibility [7]: (1) whether the user is

• Amila Silva, Ling Luo, Shanika Karunasekera, and Christopher Leckie are with the School of Computing and Information Systems, The University of Melbourne, Australia, VIC, 3010. E-mail: {amila.silva@student., ling.luo@, karus@, caleckie@}unimelb.edu.au

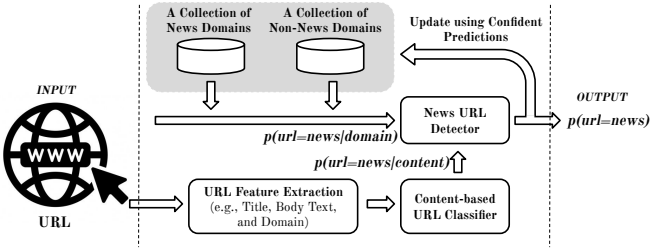

Fig. 1. Overview of the proposed pipeline to detect news URLs

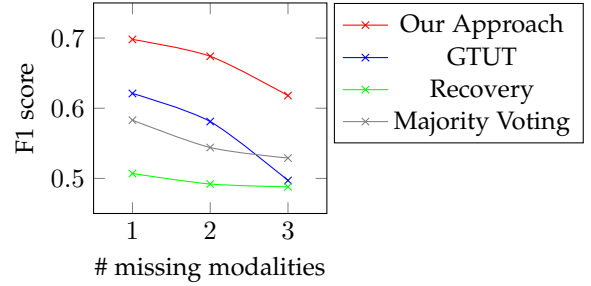

Fig. 2. F1-score of the different models for CoAID with different number of missing modalities

verified; (2) the number of followers; (3) the number of friends; (4) the number of lists; (5) the number of favourites; (6) the number of tweets; (7) the number of mentions; (8) the number of hashtags; (9) the sentiment score of the user tweet's content computed using a fine-tuned version of RoBERTa for sentiment classification[1]; and (10) the user account timestamp in months. Each feature is normalized as it helps to stabilize the subsequent learning procedure. The features of $v_\star^u$ are randomly initialized.

Based on the knowledge in the engagement-based network of a record $G^u$, we propose a mapping function $f^u : G^u \to z^u$ that learns a low-dimensional representation for the corresponding news record in a self-supervised manner.

### 2.0.1 Detecting News URLs

In the third step of *backward news dataset construction*, URLs should be classified as news URLs or non-news URLs. This could be done using two lists of news URL domains and non-news URL domains. However, this naive approach is unable to classify media-outlets that are not covered by the predefined lists. Thus, following [8], we explore whether it is possible to detect news URLs by looking at the various content features of URLs. For our study, we collect a dataset, which consists of 10,000 URLs from 30 widely known news media-outlets (e.g., *bbc.com, 9news.com, wsj.com*) and 10,000 URLs from 15 non-news media-outlets (e.g., *facebook.com, spotify.com, ebay.com*). Each URL in the dataset is represented using 5 features: (1) the number of words in the title; (2) the number of words in the body text; (3) the number of sentences in the body text; (4) the number of URLs in the webpage; and (5) the length of the URL. A Decision Tree classifier is trained using these features to classify news URLs and non-news URLs. This content-based classifier yields 97.6% in 5-fold cross-validation accuracy, which verifies the ability to detect news URLs using content features.

With the help of the content-based URL classifier, we propose a pipeline (see Figure 1) to classify an unseen URL either as news (labeled as 1) or non-news (labeled as 0). For a given URL, if its media-outlet is covered from the predefined lists of news and non-news media-outlets, our pipeline classifies the URL by merely relying on the predefined lists of media-outlets. If not, the URL is classified using the aforementioned content-based URL classifier. Also, if the confidence of the predictions for the URL as a news (non-news) URL is $\geq 95\%$ ($\leq 5\%$), the media-outlet of the URL

[1]. https://huggingface.co/cardiffnlp/twitter-roberta-base-sentiment

is added to the predefined list of news media-outlets (non-news media-outlets). This approach continuously improves the coverage of the predefined lists of media-outlet names in our pipeline, thus, increasing the speed of the pipeline. We adopt this pipeline in the third step of *backward news dataset construction* to detect news URLs. This pipeline is publicly available as a web application via https://is-a-news.herokuapp.com/.

## 3 BASELINES

The proposed framework in this work is compared against the following baselines:

- Majority Voting learns the label from multi-modal data as the majority vote from the uni-modality model from each weak source.
- UFNDA [9] adopts an autoencoder, which is learnt to return high reconstruction errors for fake news and low errors for real news.
- TruthFinder [10] iteratively calculates the truth estimation of each news based on the conflicting relationships among the verified users' tweets in an unsupervised manner.
- Recovery [11] adopts a source-credibility based approach to predict the labels for a set of seed nodes. This approach assumes that all the news records coming from reliable sources are always real and the news records coming from all unreliable sources are always fake, which is not always true. To propagate the labels to the other nodes, a textual-similarity based knowledge graph was adopted as in the proposed source credibility-based embedding function.
- UFD [12] adopts a graphical framework to identify fake news by modelling the conditional dependencies among the truths of news, the users' opinions, and the users' credibility. This baseline requires user replies for news records. Since CoAID does not include that attribute, we report the performance of this baseline only for PolitiFact and GossipCop.
- GTUT [13] adopts a graph-based label propagation mechanism to propagate labels via the propagation networks of users. To get the labels for the seed nodes with this approach, we adopt the weak labelling approach in [11].

# 4 RESULTS FOR FAKE NEWS DETECTION WITH MISSING MODALITIES

Under this experimental setting, $z$ number of modalities are randomly removed from each test instance and make predictions using different models with the available modalities. If a model cannot make a prediction using the available modalities, the label is randomly predicted. For example, if the available modalities for an instance are $P$ and $U$, Recovery cannot be used as it can exploit only $S$ and $T$ modalities. The results are shown in Figure 2. As can be seen, there is a significant performance drop for GTUT and Recovery when the number of missing modalities increase. This may be mainly due to the inability of these baselines to exploit different modalities. Our approach and Majority Voting show a relatively slight performance drop. This observation further verifies the importance of multimodal co-learning in $(UMD)^2$ when understanding multimodal environments with partially available modalities.

## REFERENCES

[1] Y. Liu, M. Ott, N. Goyal, J. Du, M. Joshi, D. Chen, O. Levy, M. Lewis, L. Zettlemoyer, and V. Stoyanov, "Roberta: A robustly optimized bert pretraining approach," *arXiv e-prints*, p. arXiv:1907.11692, 2019.
[2] S. M. Mohammad and P. D. Turney, "Nrc emotion lexicon," *National Research Council, Canada*, vol. 2, 2013.
[3] J. W. Pennebaker, R. L. Boyd, K. Jordan, and K. Blackburn, "The development and psychometric properties of LIWC2015," Tech. Rep., 2015. [Online]. Available: https://repositories.lib.utexas.edu/handle/2152/31333
[4] G. H. Mc Laughlin, "Smog grading-a new readability formula," *Journal of reading*, vol. 12, no. 8, pp. 639–646, 1969.
[5] J. Graham, J. Haidt, and B. A. Nosek, "Liberals and conservatives rely on different sets of moral foundations." *Journal of personality and social psychology*, vol. 96, no. 5, p. 1029, 2009.
[6] A. Chakraborty, B. Paranjape, S. Kakarla, and N. Ganguly, "Stop clickbait: Detecting and preventing clickbaits in online news media," in *Proc. of ASONAM*, 2016.
[7] A. Silva, Y. Han, L. Luo, S. Karunasekera, and C. Leckie, "Embedding partial propagation network for fake news early detection." in *Proc. of MAISON*, 2020.
[8] C. Arya and S. K. Dwivedi, "News web page classification using url content and structure attributes," in *Proc. of NGCT*, 2016.
[9] D. Li, H. Guo, Z. Wang, and Z. Zheng, "Unsupervised fake news detection based on autoencoder," *IEEE Access*, vol. 9, pp. 29 356–29 365, 2021.
[10] X. Yin, J. Han, and S. Y. Philip, "Truth discovery with multiple conflicting information providers on the web," *IEEE transactions on knowledge and data engineering*, 2008.
[11] X. Zhou, A. Mulay, E. Ferrara, and R. Zafarani, "Recovery: A multimodal repository for covid-19 news credibility research," in *Proc. of CIKM*, 2020.
[12] S. Yang, K. Shu, S. Wang, R. Gu, F. Wu, and H. Liu, "Unsupervised fake news detection on social media: A generative approach," in *Proc. of AAAI*, 2019.
[13] S. C. R. Gangireddy, C. Long, and T. Chakraborty, "Unsupervised fake news detection: A graph-based approach," in *Proc. of HyperText*, 2020.


\end{document}

%% file: histogram_3.pgf
%% Creator: Matplotlib, PGF backend
%%
%% To include the figure in your LaTeX document, write
%%   \input{<filename>.pgf}
%%
%% Make sure the required packages are loaded in your preamble
%%   \usepackage{pgf}
%%
%% Figures using additional raster images can only be included by \input if
%% they are in the same directory as the main LaTeX file. For loading figures
%% from other directories you can use the `import` package
%% and then include the figures with
%%   \import{<path to file>}{<filename>.pgf}
%%
%% Matplotlib used the following preamble
%%
\begingroup%
\makeatletter%
\begin{pgfpicture}%
\pgfpathrectangle{\pgfpointorigin}{\pgfqpoint{4.000000in}{2.500000in}}%
\pgfusepath{use as bounding box, clip}%
\begin{pgfscope}%
\pgfsetbuttcap%
\pgfsetmiterjoin%
\definecolor{currentfill}{rgb}{1.000000,1.000000,1.000000}%
\pgfsetfillcolor{currentfill}%
\pgfsetlinewidth{0.000000pt}%
\definecolor{currentstroke}{rgb}{1.000000,1.000000,1.000000}%
\pgfsetstrokecolor{currentstroke}%
\pgfsetdash{}{0pt}%
\pgfpathmoveto{\pgfqpoint{0.000000in}{0.000000in}}%
\pgfpathlineto{\pgfqpoint{4.000000in}{0.000000in}}%
\pgfpathlineto{\pgfqpoint{4.000000in}{2.500000in}}%
\pgfpathlineto{\pgfqpoint{0.000000in}{2.500000in}}%
\pgfpathclose%
\pgfusepath{fill}%
\end{pgfscope}%
\begin{pgfscope}%
\pgfsetbuttcap%
\pgfsetmiterjoin%
\definecolor{currentfill}{rgb}{1.000000,1.000000,1.000000}%
\pgfsetfillcolor{currentfill}%
\pgfsetlinewidth{0.000000pt}%
\definecolor{currentstroke}{rgb}{0.000000,0.000000,0.000000}%
\pgfsetstrokecolor{currentstroke}%
\pgfsetstrokeopacity{0.000000}%
\pgfsetdash{}{0pt}%
\pgfpathmoveto{\pgfqpoint{0.500000in}{0.312500in}}%
\pgfpathlineto{\pgfqpoint{3.600000in}{0.312500in}}%
\pgfpathlineto{\pgfqpoint{3.600000in}{2.200000in}}%
\pgfpathlineto{\pgfqpoint{0.500000in}{2.200000in}}%
\pgfpathclose%
\pgfusepath{fill}%
\end{pgfscope}%
\begin{pgfscope}%
\pgfsetbuttcap%
\pgfsetroundjoin%
\definecolor{currentfill}{rgb}{0.000000,0.000000,0.000000}%
\pgfsetfillcolor{currentfill}%
\pgfsetlinewidth{0.803000pt}%
\definecolor{currentstroke}{rgb}{0.000000,0.000000,0.000000}%
\pgfsetstrokecolor{currentstroke}%
\pgfsetdash{}{0pt}%
\pgfsys@defobject{currentmarker}{\pgfqpoint{0.000000in}{-0.048611in}}{\pgfqpoint{0.000000in}{0.000000in}}{%
\pgfpathmoveto{\pgfqpoint{0.000000in}{0.000000in}}%
\pgfpathlineto{\pgfqpoint{0.000000in}{-0.048611in}}%
\pgfusepath{stroke,fill}%
}%
\begin{pgfscope}%
\pgfsys@transformshift{0.640909in}{0.312500in}%
\pgfsys@useobject{currentmarker}{}%
\end{pgfscope}%
\end{pgfscope}%
\begin{pgfscope}%
\definecolor{textcolor}{rgb}{0.000000,0.000000,0.000000}%
\pgfsetstrokecolor{textcolor}%
\pgfsetfillcolor{textcolor}%
\pgftext[x=0.640909in,y=0.215278in,,top]{\color{textcolor}\rmfamily\fontsize{15.000000}{18.000000}\selectfont \(\displaystyle 0\)}%
\end{pgfscope}%
\begin{pgfscope}%
\pgfsetbuttcap%
\pgfsetroundjoin%
\definecolor{currentfill}{rgb}{0.000000,0.000000,0.000000}%
\pgfsetfillcolor{currentfill}%
\pgfsetlinewidth{0.803000pt}%
\definecolor{currentstroke}{rgb}{0.000000,0.000000,0.000000}%
\pgfsetstrokecolor{currentstroke}%
\pgfsetdash{}{0pt}%
\pgfsys@defobject{currentmarker}{\pgfqpoint{0.000000in}{-0.048611in}}{\pgfqpoint{0.000000in}{0.000000in}}{%
\pgfpathmoveto{\pgfqpoint{0.000000in}{0.000000in}}%
\pgfpathlineto{\pgfqpoint{0.000000in}{-0.048611in}}%
\pgfusepath{stroke,fill}%
}%
\begin{pgfscope}%
\pgfsys@transformshift{1.345455in}{0.312500in}%
\pgfsys@useobject{currentmarker}{}%
\end{pgfscope}%
\end{pgfscope}%
\begin{pgfscope}%
\definecolor{textcolor}{rgb}{0.000000,0.000000,0.000000}%
\pgfsetstrokecolor{textcolor}%
\pgfsetfillcolor{textcolor}%
\pgftext[x=1.345455in,y=0.215278in,,top]{\color{textcolor}\rmfamily\fontsize{15.000000}{18.000000}\selectfont \(\displaystyle 5\)}%
\end{pgfscope}%
\begin{pgfscope}%
\pgfsetbuttcap%
\pgfsetroundjoin%
\definecolor{currentfill}{rgb}{0.000000,0.000000,0.000000}%
\pgfsetfillcolor{currentfill}%
\pgfsetlinewidth{0.803000pt}%
\definecolor{currentstroke}{rgb}{0.000000,0.000000,0.000000}%
\pgfsetstrokecolor{currentstroke}%
\pgfsetdash{}{0pt}%
\pgfsys@defobject{currentmarker}{\pgfqpoint{0.000000in}{-0.048611in}}{\pgfqpoint{0.000000in}{0.000000in}}{%
\pgfpathmoveto{\pgfqpoint{0.000000in}{0.000000in}}%
\pgfpathlineto{\pgfqpoint{0.000000in}{-0.048611in}}%
\pgfusepath{stroke,fill}%
}%
\begin{pgfscope}%
\pgfsys@transformshift{2.050000in}{0.312500in}%
\pgfsys@useobject{currentmarker}{}%
\end{pgfscope}%
\end{pgfscope}%
\begin{pgfscope}%
\definecolor{textcolor}{rgb}{0.000000,0.000000,0.000000}%
\pgfsetstrokecolor{textcolor}%
\pgfsetfillcolor{textcolor}%
\pgftext[x=2.050000in,y=0.215278in,,top]{\color{textcolor}\rmfamily\fontsize{15.000000}{18.000000}\selectfont \(\displaystyle 10\)}%
\end{pgfscope}%
\begin{pgfscope}%
\pgfsetbuttcap%
\pgfsetroundjoin%
\definecolor{currentfill}{rgb}{0.000000,0.000000,0.000000}%
\pgfsetfillcolor{currentfill}%
\pgfsetlinewidth{0.803000pt}%
\definecolor{currentstroke}{rgb}{0.000000,0.000000,0.000000}%
\pgfsetstrokecolor{currentstroke}%
\pgfsetdash{}{0pt}%
\pgfsys@defobject{currentmarker}{\pgfqpoint{0.000000in}{-0.048611in}}{\pgfqpoint{0.000000in}{0.000000in}}{%
\pgfpathmoveto{\pgfqpoint{0.000000in}{0.000000in}}%
\pgfpathlineto{\pgfqpoint{0.000000in}{-0.048611in}}%
\pgfusepath{stroke,fill}%
}%
\begin{pgfscope}%
\pgfsys@transformshift{2.754545in}{0.312500in}%
\pgfsys@useobject{currentmarker}{}%
\end{pgfscope}%
\end{pgfscope}%
\begin{pgfscope}%
\definecolor{textcolor}{rgb}{0.000000,0.000000,0.000000}%
\pgfsetstrokecolor{textcolor}%
\pgfsetfillcolor{textcolor}%
\pgftext[x=2.754545in,y=0.215278in,,top]{\color{textcolor}\rmfamily\fontsize{15.000000}{18.000000}\selectfont \(\displaystyle 15\)}%
\end{pgfscope}%
\begin{pgfscope}%
\pgfsetbuttcap%
\pgfsetroundjoin%
\definecolor{currentfill}{rgb}{0.000000,0.000000,0.000000}%
\pgfsetfillcolor{currentfill}%
\pgfsetlinewidth{0.803000pt}%
\definecolor{currentstroke}{rgb}{0.000000,0.000000,0.000000}%
\pgfsetstrokecolor{currentstroke}%
\pgfsetdash{}{0pt}%
\pgfsys@defobject{currentmarker}{\pgfqpoint{0.000000in}{-0.048611in}}{\pgfqpoint{0.000000in}{0.000000in}}{%
\pgfpathmoveto{\pgfqpoint{0.000000in}{0.000000in}}%
\pgfpathlineto{\pgfqpoint{0.000000in}{-0.048611in}}%
\pgfusepath{stroke,fill}%
}%
\begin{pgfscope}%
\pgfsys@transformshift{3.459091in}{0.312500in}%
\pgfsys@useobject{currentmarker}{}%
\end{pgfscope}%
\end{pgfscope}%
\begin{pgfscope}%
\definecolor{textcolor}{rgb}{0.000000,0.000000,0.000000}%
\pgfsetstrokecolor{textcolor}%
\pgfsetfillcolor{textcolor}%
\pgftext[x=3.459091in,y=0.215278in,,top]{\color{textcolor}\rmfamily\fontsize{15.000000}{18.000000}\selectfont \(\displaystyle 20\)}%
\end{pgfscope}%
\begin{pgfscope}%
\definecolor{textcolor}{rgb}{0.000000,0.000000,0.000000}%
\pgfsetstrokecolor{textcolor}%
\pgfsetfillcolor{textcolor}%
\pgftext[x=2.050000in,y=-0.018055in,,top]{\color{textcolor}\rmfamily\fontsize{15.000000}{18.000000}\selectfont k}%
\end{pgfscope}%
\begin{pgfscope}%
\pgfsetbuttcap%
\pgfsetroundjoin%
\definecolor{currentfill}{rgb}{0.000000,0.000000,0.000000}%
\pgfsetfillcolor{currentfill}%
\pgfsetlinewidth{0.803000pt}%
\definecolor{currentstroke}{rgb}{0.000000,0.000000,0.000000}%
\pgfsetstrokecolor{currentstroke}%
\pgfsetdash{}{0pt}%
\pgfsys@defobject{currentmarker}{\pgfqpoint{-0.048611in}{0.000000in}}{\pgfqpoint{0.000000in}{0.000000in}}{%
\pgfpathmoveto{\pgfqpoint{0.000000in}{0.000000in}}%
\pgfpathlineto{\pgfqpoint{-0.048611in}{0.000000in}}%
\pgfusepath{stroke,fill}%
}%
\begin{pgfscope}%
\pgfsys@transformshift{0.500000in}{0.604205in}%
\pgfsys@useobject{currentmarker}{}%
\end{pgfscope}%
\end{pgfscope}%
\begin{pgfscope}%
\definecolor{textcolor}{rgb}{0.000000,0.000000,0.000000}%
\pgfsetstrokecolor{textcolor}%
\pgfsetfillcolor{textcolor}%
\pgftext[x=0.054634in,y=0.534760in,left,base]{\color{textcolor}\rmfamily\fontsize{15.000000}{18.000000}\selectfont \(\displaystyle 0.72\)}%
\end{pgfscope}%
\begin{pgfscope}%
\pgfsetbuttcap%
\pgfsetroundjoin%
\definecolor{currentfill}{rgb}{0.000000,0.000000,0.000000}%
\pgfsetfillcolor{currentfill}%
\pgfsetlinewidth{0.803000pt}%
\definecolor{currentstroke}{rgb}{0.000000,0.000000,0.000000}%
\pgfsetstrokecolor{currentstroke}%
\pgfsetdash{}{0pt}%
\pgfsys@defobject{currentmarker}{\pgfqpoint{-0.048611in}{0.000000in}}{\pgfqpoint{0.000000in}{0.000000in}}{%
\pgfpathmoveto{\pgfqpoint{0.000000in}{0.000000in}}%
\pgfpathlineto{\pgfqpoint{-0.048611in}{0.000000in}}%
\pgfusepath{stroke,fill}%
}%
\begin{pgfscope}%
\pgfsys@transformshift{0.500000in}{0.947386in}%
\pgfsys@useobject{currentmarker}{}%
\end{pgfscope}%
\end{pgfscope}%
\begin{pgfscope}%
\definecolor{textcolor}{rgb}{0.000000,0.000000,0.000000}%
\pgfsetstrokecolor{textcolor}%
\pgfsetfillcolor{textcolor}%
\pgftext[x=0.054634in,y=0.877942in,left,base]{\color{textcolor}\rmfamily\fontsize{15.000000}{18.000000}\selectfont \(\displaystyle 0.74\)}%
\end{pgfscope}%
\begin{pgfscope}%
\pgfsetbuttcap%
\pgfsetroundjoin%
\definecolor{currentfill}{rgb}{0.000000,0.000000,0.000000}%
\pgfsetfillcolor{currentfill}%
\pgfsetlinewidth{0.803000pt}%
\definecolor{currentstroke}{rgb}{0.000000,0.000000,0.000000}%
\pgfsetstrokecolor{currentstroke}%
\pgfsetdash{}{0pt}%
\pgfsys@defobject{currentmarker}{\pgfqpoint{-0.048611in}{0.000000in}}{\pgfqpoint{0.000000in}{0.000000in}}{%
\pgfpathmoveto{\pgfqpoint{0.000000in}{0.000000in}}%
\pgfpathlineto{\pgfqpoint{-0.048611in}{0.000000in}}%
\pgfusepath{stroke,fill}%
}%
\begin{pgfscope}%
\pgfsys@transformshift{0.500000in}{1.290568in}%
\pgfsys@useobject{currentmarker}{}%
\end{pgfscope}%
\end{pgfscope}%
\begin{pgfscope}%
\definecolor{textcolor}{rgb}{0.000000,0.000000,0.000000}%
\pgfsetstrokecolor{textcolor}%
\pgfsetfillcolor{textcolor}%
\pgftext[x=0.054634in,y=1.221124in,left,base]{\color{textcolor}\rmfamily\fontsize{15.000000}{18.000000}\selectfont \(\displaystyle 0.76\)}%
\end{pgfscope}%
\begin{pgfscope}%
\pgfsetbuttcap%
\pgfsetroundjoin%
\definecolor{currentfill}{rgb}{0.000000,0.000000,0.000000}%
\pgfsetfillcolor{currentfill}%
\pgfsetlinewidth{0.803000pt}%
\definecolor{currentstroke}{rgb}{0.000000,0.000000,0.000000}%
\pgfsetstrokecolor{currentstroke}%
\pgfsetdash{}{0pt}%
\pgfsys@defobject{currentmarker}{\pgfqpoint{-0.048611in}{0.000000in}}{\pgfqpoint{0.000000in}{0.000000in}}{%
\pgfpathmoveto{\pgfqpoint{0.000000in}{0.000000in}}%
\pgfpathlineto{\pgfqpoint{-0.048611in}{0.000000in}}%
\pgfusepath{stroke,fill}%
}%
\begin{pgfscope}%
\pgfsys@transformshift{0.500000in}{1.633750in}%
\pgfsys@useobject{currentmarker}{}%
\end{pgfscope}%
\end{pgfscope}%
\begin{pgfscope}%
\definecolor{textcolor}{rgb}{0.000000,0.000000,0.000000}%
\pgfsetstrokecolor{textcolor}%
\pgfsetfillcolor{textcolor}%
\pgftext[x=0.054634in,y=1.564306in,left,base]{\color{textcolor}\rmfamily\fontsize{15.000000}{18.000000}\selectfont \(\displaystyle 0.78\)}%
\end{pgfscope}%
\begin{pgfscope}%
\pgfsetbuttcap%
\pgfsetroundjoin%
\definecolor{currentfill}{rgb}{0.000000,0.000000,0.000000}%
\pgfsetfillcolor{currentfill}%
\pgfsetlinewidth{0.803000pt}%
\definecolor{currentstroke}{rgb}{0.000000,0.000000,0.000000}%
\pgfsetstrokecolor{currentstroke}%
\pgfsetdash{}{0pt}%
\pgfsys@defobject{currentmarker}{\pgfqpoint{-0.048611in}{0.000000in}}{\pgfqpoint{0.000000in}{0.000000in}}{%
\pgfpathmoveto{\pgfqpoint{0.000000in}{0.000000in}}%
\pgfpathlineto{\pgfqpoint{-0.048611in}{0.000000in}}%
\pgfusepath{stroke,fill}%
}%
\begin{pgfscope}%
\pgfsys@transformshift{0.500000in}{1.976932in}%
\pgfsys@useobject{currentmarker}{}%
\end{pgfscope}%
\end{pgfscope}%
\begin{pgfscope}%
\definecolor{textcolor}{rgb}{0.000000,0.000000,0.000000}%
\pgfsetstrokecolor{textcolor}%
\pgfsetfillcolor{textcolor}%
\pgftext[x=0.054634in,y=1.907488in,left,base]{\color{textcolor}\rmfamily\fontsize{15.000000}{18.000000}\selectfont \(\displaystyle 0.80\)}%
\end{pgfscope}%
\begin{pgfscope}%
\definecolor{textcolor}{rgb}{0.000000,0.000000,0.000000}%
\pgfsetstrokecolor{textcolor}%
\pgfsetfillcolor{textcolor}%
\pgftext[x=-0.000922in,y=1.256250in,,bottom,rotate=90.000000]{\color{textcolor}\rmfamily\fontsize{15.000000}{18.000000}\selectfont F1-score}%
\end{pgfscope}%
\begin{pgfscope}%
\pgfpathrectangle{\pgfqpoint{0.500000in}{0.312500in}}{\pgfqpoint{3.100000in}{1.887500in}}%
\pgfusepath{clip}%
\pgfsetrectcap%
\pgfsetroundjoin%
\pgfsetlinewidth{1.505625pt}%
\definecolor{currentstroke}{rgb}{1.000000,0.000000,0.000000}%
\pgfsetstrokecolor{currentstroke}%
\pgfsetdash{}{0pt}%
\pgfpathmoveto{\pgfqpoint{0.640909in}{0.398295in}}%
\pgfpathlineto{\pgfqpoint{1.345455in}{1.307727in}}%
\pgfpathlineto{\pgfqpoint{2.050000in}{1.668068in}}%
\pgfpathlineto{\pgfqpoint{2.754545in}{1.771023in}}%
\pgfpathlineto{\pgfqpoint{3.459091in}{1.753864in}}%
\pgfusepath{stroke}%
\end{pgfscope}%
\begin{pgfscope}%
\pgfpathrectangle{\pgfqpoint{0.500000in}{0.312500in}}{\pgfqpoint{3.100000in}{1.887500in}}%
\pgfusepath{clip}%
\pgfsetbuttcap%
\pgfsetroundjoin%
\definecolor{currentfill}{rgb}{1.000000,0.000000,0.000000}%
\pgfsetfillcolor{currentfill}%
\pgfsetlinewidth{1.003750pt}%
\definecolor{currentstroke}{rgb}{1.000000,0.000000,0.000000}%
\pgfsetstrokecolor{currentstroke}%
\pgfsetdash{}{0pt}%
\pgfsys@defobject{currentmarker}{\pgfqpoint{-0.041667in}{-0.041667in}}{\pgfqpoint{0.041667in}{0.041667in}}{%
\pgfpathmoveto{\pgfqpoint{0.000000in}{-0.041667in}}%
\pgfpathcurveto{\pgfqpoint{0.011050in}{-0.041667in}}{\pgfqpoint{0.021649in}{-0.037276in}}{\pgfqpoint{0.029463in}{-0.029463in}}%
\pgfpathcurveto{\pgfqpoint{0.037276in}{-0.021649in}}{\pgfqpoint{0.041667in}{-0.011050in}}{\pgfqpoint{0.041667in}{0.000000in}}%
\pgfpathcurveto{\pgfqpoint{0.041667in}{0.011050in}}{\pgfqpoint{0.037276in}{0.021649in}}{\pgfqpoint{0.029463in}{0.029463in}}%
\pgfpathcurveto{\pgfqpoint{0.021649in}{0.037276in}}{\pgfqpoint{0.011050in}{0.041667in}}{\pgfqpoint{0.000000in}{0.041667in}}%
\pgfpathcurveto{\pgfqpoint{-0.011050in}{0.041667in}}{\pgfqpoint{-0.021649in}{0.037276in}}{\pgfqpoint{-0.029463in}{0.029463in}}%
\pgfpathcurveto{\pgfqpoint{-0.037276in}{0.021649in}}{\pgfqpoint{-0.041667in}{0.011050in}}{\pgfqpoint{-0.041667in}{0.000000in}}%
\pgfpathcurveto{\pgfqpoint{-0.041667in}{-0.011050in}}{\pgfqpoint{-0.037276in}{-0.021649in}}{\pgfqpoint{-0.029463in}{-0.029463in}}%
\pgfpathcurveto{\pgfqpoint{-0.021649in}{-0.037276in}}{\pgfqpoint{-0.011050in}{-0.041667in}}{\pgfqpoint{0.000000in}{-0.041667in}}%
\pgfpathclose%
\pgfusepath{stroke,fill}%
}%
\begin{pgfscope}%
\pgfsys@transformshift{0.640909in}{0.398295in}%
\pgfsys@useobject{currentmarker}{}%
\end{pgfscope}%
\begin{pgfscope}%
\pgfsys@transformshift{1.345455in}{1.307727in}%
\pgfsys@useobject{currentmarker}{}%
\end{pgfscope}%
\begin{pgfscope}%
\pgfsys@transformshift{2.050000in}{1.668068in}%
\pgfsys@useobject{currentmarker}{}%
\end{pgfscope}%
\begin{pgfscope}%
\pgfsys@transformshift{2.754545in}{1.771023in}%
\pgfsys@useobject{currentmarker}{}%
\end{pgfscope}%
\begin{pgfscope}%
\pgfsys@transformshift{3.459091in}{1.753864in}%
\pgfsys@useobject{currentmarker}{}%
\end{pgfscope}%
\end{pgfscope}%
\begin{pgfscope}%
\pgfpathrectangle{\pgfqpoint{0.500000in}{0.312500in}}{\pgfqpoint{3.100000in}{1.887500in}}%
\pgfusepath{clip}%
\pgfsetrectcap%
\pgfsetroundjoin%
\pgfsetlinewidth{1.505625pt}%
\definecolor{currentstroke}{rgb}{0.000000,0.000000,1.000000}%
\pgfsetstrokecolor{currentstroke}%
\pgfsetdash{}{0pt}%
\pgfpathmoveto{\pgfqpoint{0.640909in}{1.307727in}}%
\pgfpathlineto{\pgfqpoint{1.345455in}{1.736705in}}%
\pgfpathlineto{\pgfqpoint{2.050000in}{1.925455in}}%
\pgfpathlineto{\pgfqpoint{2.754545in}{2.062727in}}%
\pgfpathlineto{\pgfqpoint{3.459091in}{2.114205in}}%
\pgfusepath{stroke}%
\end{pgfscope}%
\begin{pgfscope}%
\pgfpathrectangle{\pgfqpoint{0.500000in}{0.312500in}}{\pgfqpoint{3.100000in}{1.887500in}}%
\pgfusepath{clip}%
\pgfsetbuttcap%
\pgfsetmiterjoin%
\definecolor{currentfill}{rgb}{0.000000,0.000000,1.000000}%
\pgfsetfillcolor{currentfill}%
\pgfsetlinewidth{1.003750pt}%
\definecolor{currentstroke}{rgb}{0.000000,0.000000,1.000000}%
\pgfsetstrokecolor{currentstroke}%
\pgfsetdash{}{0pt}%
\pgfsys@defobject{currentmarker}{\pgfqpoint{-0.041667in}{-0.041667in}}{\pgfqpoint{0.041667in}{0.041667in}}{%
\pgfpathmoveto{\pgfqpoint{-0.041667in}{-0.041667in}}%
\pgfpathlineto{\pgfqpoint{0.041667in}{-0.041667in}}%
\pgfpathlineto{\pgfqpoint{0.041667in}{0.041667in}}%
\pgfpathlineto{\pgfqpoint{-0.041667in}{0.041667in}}%
\pgfpathclose%
\pgfusepath{stroke,fill}%
}%
\begin{pgfscope}%
\pgfsys@transformshift{0.640909in}{1.307727in}%
\pgfsys@useobject{currentmarker}{}%
\end{pgfscope}%
\begin{pgfscope}%
\pgfsys@transformshift{1.345455in}{1.736705in}%
\pgfsys@useobject{currentmarker}{}%
\end{pgfscope}%
\begin{pgfscope}%
\pgfsys@transformshift{2.050000in}{1.925455in}%
\pgfsys@useobject{currentmarker}{}%
\end{pgfscope}%
\begin{pgfscope}%
\pgfsys@transformshift{2.754545in}{2.062727in}%
\pgfsys@useobject{currentmarker}{}%
\end{pgfscope}%
\begin{pgfscope}%
\pgfsys@transformshift{3.459091in}{2.114205in}%
\pgfsys@useobject{currentmarker}{}%
\end{pgfscope}%
\end{pgfscope}%
\begin{pgfscope}%
\pgfpathrectangle{\pgfqpoint{0.500000in}{0.312500in}}{\pgfqpoint{3.100000in}{1.887500in}}%
\pgfusepath{clip}%
\pgfsetrectcap%
\pgfsetroundjoin%
\pgfsetlinewidth{1.505625pt}%
\definecolor{currentstroke}{rgb}{0.000000,0.500000,0.000000}%
\pgfsetstrokecolor{currentstroke}%
\pgfsetdash{}{0pt}%
\pgfpathmoveto{\pgfqpoint{0.640909in}{1.685227in}}%
\pgfpathlineto{\pgfqpoint{1.345455in}{1.822500in}}%
\pgfpathlineto{\pgfqpoint{2.050000in}{1.925455in}}%
\pgfpathlineto{\pgfqpoint{2.754545in}{1.891136in}}%
\pgfpathlineto{\pgfqpoint{3.459091in}{1.959773in}}%
\pgfusepath{stroke}%
\end{pgfscope}%
\begin{pgfscope}%
\pgfpathrectangle{\pgfqpoint{0.500000in}{0.312500in}}{\pgfqpoint{3.100000in}{1.887500in}}%
\pgfusepath{clip}%
\pgfsetbuttcap%
\pgfsetmiterjoin%
\definecolor{currentfill}{rgb}{0.000000,0.500000,0.000000}%
\pgfsetfillcolor{currentfill}%
\pgfsetlinewidth{1.003750pt}%
\definecolor{currentstroke}{rgb}{0.000000,0.500000,0.000000}%
\pgfsetstrokecolor{currentstroke}%
\pgfsetdash{}{0pt}%
\pgfsys@defobject{currentmarker}{\pgfqpoint{-0.041667in}{-0.041667in}}{\pgfqpoint{0.041667in}{0.041667in}}{%
\pgfpathmoveto{\pgfqpoint{0.000000in}{0.041667in}}%
\pgfpathlineto{\pgfqpoint{-0.041667in}{-0.041667in}}%
\pgfpathlineto{\pgfqpoint{0.041667in}{-0.041667in}}%
\pgfpathclose%
\pgfusepath{stroke,fill}%
}%
\begin{pgfscope}%
\pgfsys@transformshift{0.640909in}{1.685227in}%
\pgfsys@useobject{currentmarker}{}%
\end{pgfscope}%
\begin{pgfscope}%
\pgfsys@transformshift{1.345455in}{1.822500in}%
\pgfsys@useobject{currentmarker}{}%
\end{pgfscope}%
\begin{pgfscope}%
\pgfsys@transformshift{2.050000in}{1.925455in}%
\pgfsys@useobject{currentmarker}{}%
\end{pgfscope}%
\begin{pgfscope}%
\pgfsys@transformshift{2.754545in}{1.891136in}%
\pgfsys@useobject{currentmarker}{}%
\end{pgfscope}%
\begin{pgfscope}%
\pgfsys@transformshift{3.459091in}{1.959773in}%
\pgfsys@useobject{currentmarker}{}%
\end{pgfscope}%
\end{pgfscope}%
\begin{pgfscope}%
\pgfsetrectcap%
\pgfsetmiterjoin%
\pgfsetlinewidth{0.803000pt}%
\definecolor{currentstroke}{rgb}{0.000000,0.000000,0.000000}%
\pgfsetstrokecolor{currentstroke}%
\pgfsetdash{}{0pt}%
\pgfpathmoveto{\pgfqpoint{0.500000in}{0.312500in}}%
\pgfpathlineto{\pgfqpoint{0.500000in}{2.200000in}}%
\pgfusepath{stroke}%
\end{pgfscope}%
\begin{pgfscope}%
\pgfsetrectcap%
\pgfsetmiterjoin%
\pgfsetlinewidth{0.803000pt}%
\definecolor{currentstroke}{rgb}{0.000000,0.000000,0.000000}%
\pgfsetstrokecolor{currentstroke}%
\pgfsetdash{}{0pt}%
\pgfpathmoveto{\pgfqpoint{3.600000in}{0.312500in}}%
\pgfpathlineto{\pgfqpoint{3.600000in}{2.200000in}}%
\pgfusepath{stroke}%
\end{pgfscope}%
\begin{pgfscope}%
\pgfsetrectcap%
\pgfsetmiterjoin%
\pgfsetlinewidth{0.803000pt}%
\definecolor{currentstroke}{rgb}{0.000000,0.000000,0.000000}%
\pgfsetstrokecolor{currentstroke}%
\pgfsetdash{}{0pt}%
\pgfpathmoveto{\pgfqpoint{0.500000in}{0.312500in}}%
\pgfpathlineto{\pgfqpoint{3.600000in}{0.312500in}}%
\pgfusepath{stroke}%
\end{pgfscope}%
\begin{pgfscope}%
\pgfsetrectcap%
\pgfsetmiterjoin%
\pgfsetlinewidth{0.803000pt}%
\definecolor{currentstroke}{rgb}{0.000000,0.000000,0.000000}%
\pgfsetstrokecolor{currentstroke}%
\pgfsetdash{}{0pt}%
\pgfpathmoveto{\pgfqpoint{0.500000in}{2.200000in}}%
\pgfpathlineto{\pgfqpoint{3.600000in}{2.200000in}}%
\pgfusepath{stroke}%
\end{pgfscope}%
\begin{pgfscope}%
\pgfsetbuttcap%
\pgfsetmiterjoin%
\definecolor{currentfill}{rgb}{1.000000,1.000000,1.000000}%
\pgfsetfillcolor{currentfill}%
\pgfsetfillopacity{0.800000}%
\pgfsetlinewidth{1.003750pt}%
\definecolor{currentstroke}{rgb}{0.800000,0.800000,0.800000}%
\pgfsetstrokecolor{currentstroke}%
\pgfsetstrokeopacity{0.800000}%
\pgfsetdash{}{0pt}%
\pgfpathmoveto{\pgfqpoint{1.870104in}{0.416667in}}%
\pgfpathlineto{\pgfqpoint{3.454167in}{0.416667in}}%
\pgfpathquadraticcurveto{\pgfqpoint{3.495833in}{0.416667in}}{\pgfqpoint{3.495833in}{0.458333in}}%
\pgfpathlineto{\pgfqpoint{3.495833in}{1.304166in}}%
\pgfpathquadraticcurveto{\pgfqpoint{3.495833in}{1.345833in}}{\pgfqpoint{3.454167in}{1.345833in}}%
\pgfpathlineto{\pgfqpoint{1.870104in}{1.345833in}}%
\pgfpathquadraticcurveto{\pgfqpoint{1.828437in}{1.345833in}}{\pgfqpoint{1.828437in}{1.304166in}}%
\pgfpathlineto{\pgfqpoint{1.828437in}{0.458333in}}%
\pgfpathquadraticcurveto{\pgfqpoint{1.828437in}{0.416667in}}{\pgfqpoint{1.870104in}{0.416667in}}%
\pgfpathclose%
\pgfusepath{stroke,fill}%
\end{pgfscope}%
\begin{pgfscope}%
\pgfsetrectcap%
\pgfsetroundjoin%
\pgfsetlinewidth{1.505625pt}%
\definecolor{currentstroke}{rgb}{1.000000,0.000000,0.000000}%
\pgfsetstrokecolor{currentstroke}%
\pgfsetdash{}{0pt}%
\pgfpathmoveto{\pgfqpoint{1.911770in}{1.189583in}}%
\pgfpathlineto{\pgfqpoint{2.328437in}{1.189583in}}%
\pgfusepath{stroke}%
\end{pgfscope}%
\begin{pgfscope}%
\pgfsetbuttcap%
\pgfsetroundjoin%
\definecolor{currentfill}{rgb}{1.000000,0.000000,0.000000}%
\pgfsetfillcolor{currentfill}%
\pgfsetlinewidth{1.003750pt}%
\definecolor{currentstroke}{rgb}{1.000000,0.000000,0.000000}%
\pgfsetstrokecolor{currentstroke}%
\pgfsetdash{}{0pt}%
\pgfsys@defobject{currentmarker}{\pgfqpoint{-0.041667in}{-0.041667in}}{\pgfqpoint{0.041667in}{0.041667in}}{%
\pgfpathmoveto{\pgfqpoint{0.000000in}{-0.041667in}}%
\pgfpathcurveto{\pgfqpoint{0.011050in}{-0.041667in}}{\pgfqpoint{0.021649in}{-0.037276in}}{\pgfqpoint{0.029463in}{-0.029463in}}%
\pgfpathcurveto{\pgfqpoint{0.037276in}{-0.021649in}}{\pgfqpoint{0.041667in}{-0.011050in}}{\pgfqpoint{0.041667in}{0.000000in}}%
\pgfpathcurveto{\pgfqpoint{0.041667in}{0.011050in}}{\pgfqpoint{0.037276in}{0.021649in}}{\pgfqpoint{0.029463in}{0.029463in}}%
\pgfpathcurveto{\pgfqpoint{0.021649in}{0.037276in}}{\pgfqpoint{0.011050in}{0.041667in}}{\pgfqpoint{0.000000in}{0.041667in}}%
\pgfpathcurveto{\pgfqpoint{-0.011050in}{0.041667in}}{\pgfqpoint{-0.021649in}{0.037276in}}{\pgfqpoint{-0.029463in}{0.029463in}}%
\pgfpathcurveto{\pgfqpoint{-0.037276in}{0.021649in}}{\pgfqpoint{-0.041667in}{0.011050in}}{\pgfqpoint{-0.041667in}{0.000000in}}%
\pgfpathcurveto{\pgfqpoint{-0.041667in}{-0.011050in}}{\pgfqpoint{-0.037276in}{-0.021649in}}{\pgfqpoint{-0.029463in}{-0.029463in}}%
\pgfpathcurveto{\pgfqpoint{-0.021649in}{-0.037276in}}{\pgfqpoint{-0.011050in}{-0.041667in}}{\pgfqpoint{0.000000in}{-0.041667in}}%
\pgfpathclose%
\pgfusepath{stroke,fill}%
}%
\begin{pgfscope}%
\pgfsys@transformshift{2.120104in}{1.189583in}%
\pgfsys@useobject{currentmarker}{}%
\end{pgfscope}%
\end{pgfscope}%
\begin{pgfscope}%
\definecolor{textcolor}{rgb}{0.000000,0.000000,0.000000}%
\pgfsetstrokecolor{textcolor}%
\pgfsetfillcolor{textcolor}%
\pgftext[x=2.495104in,y=1.116666in,left,base]{\color{textcolor}\rmfamily\fontsize{15.000000}{18.000000}\selectfont CoAID}%
\end{pgfscope}%
\begin{pgfscope}%
\pgfsetrectcap%
\pgfsetroundjoin%
\pgfsetlinewidth{1.505625pt}%
\definecolor{currentstroke}{rgb}{0.000000,0.000000,1.000000}%
\pgfsetstrokecolor{currentstroke}%
\pgfsetdash{}{0pt}%
\pgfpathmoveto{\pgfqpoint{1.911770in}{0.900694in}}%
\pgfpathlineto{\pgfqpoint{2.328437in}{0.900694in}}%
\pgfusepath{stroke}%
\end{pgfscope}%
\begin{pgfscope}%
\pgfsetbuttcap%
\pgfsetmiterjoin%
\definecolor{currentfill}{rgb}{0.000000,0.000000,1.000000}%
\pgfsetfillcolor{currentfill}%
\pgfsetlinewidth{1.003750pt}%
\definecolor{currentstroke}{rgb}{0.000000,0.000000,1.000000}%
\pgfsetstrokecolor{currentstroke}%
\pgfsetdash{}{0pt}%
\pgfsys@defobject{currentmarker}{\pgfqpoint{-0.041667in}{-0.041667in}}{\pgfqpoint{0.041667in}{0.041667in}}{%
\pgfpathmoveto{\pgfqpoint{-0.041667in}{-0.041667in}}%
\pgfpathlineto{\pgfqpoint{0.041667in}{-0.041667in}}%
\pgfpathlineto{\pgfqpoint{0.041667in}{0.041667in}}%
\pgfpathlineto{\pgfqpoint{-0.041667in}{0.041667in}}%
\pgfpathclose%
\pgfusepath{stroke,fill}%
}%
\begin{pgfscope}%
\pgfsys@transformshift{2.120104in}{0.900694in}%
\pgfsys@useobject{currentmarker}{}%
\end{pgfscope}%
\end{pgfscope}%
\begin{pgfscope}%
\definecolor{textcolor}{rgb}{0.000000,0.000000,0.000000}%
\pgfsetstrokecolor{textcolor}%
\pgfsetfillcolor{textcolor}%
\pgftext[x=2.495104in,y=0.827777in,left,base]{\color{textcolor}\rmfamily\fontsize{15.000000}{18.000000}\selectfont PolitiFact}%
\end{pgfscope}%
\begin{pgfscope}%
\pgfsetrectcap%
\pgfsetroundjoin%
\pgfsetlinewidth{1.505625pt}%
\definecolor{currentstroke}{rgb}{0.000000,0.500000,0.000000}%
\pgfsetstrokecolor{currentstroke}%
\pgfsetdash{}{0pt}%
\pgfpathmoveto{\pgfqpoint{1.911770in}{0.611805in}}%
\pgfpathlineto{\pgfqpoint{2.328437in}{0.611805in}}%
\pgfusepath{stroke}%
\end{pgfscope}%
\begin{pgfscope}%
\pgfsetbuttcap%
\pgfsetmiterjoin%
\definecolor{currentfill}{rgb}{0.000000,0.500000,0.000000}%
\pgfsetfillcolor{currentfill}%
\pgfsetlinewidth{1.003750pt}%
\definecolor{currentstroke}{rgb}{0.000000,0.500000,0.000000}%
\pgfsetstrokecolor{currentstroke}%
\pgfsetdash{}{0pt}%
\pgfsys@defobject{currentmarker}{\pgfqpoint{-0.041667in}{-0.041667in}}{\pgfqpoint{0.041667in}{0.041667in}}{%
\pgfpathmoveto{\pgfqpoint{0.000000in}{0.041667in}}%
\pgfpathlineto{\pgfqpoint{-0.041667in}{-0.041667in}}%
\pgfpathlineto{\pgfqpoint{0.041667in}{-0.041667in}}%
\pgfpathclose%
\pgfusepath{stroke,fill}%
}%
\begin{pgfscope}%
\pgfsys@transformshift{2.120104in}{0.611805in}%
\pgfsys@useobject{currentmarker}{}%
\end{pgfscope}%
\end{pgfscope}%
\begin{pgfscope}%
\definecolor{textcolor}{rgb}{0.000000,0.000000,0.000000}%
\pgfsetstrokecolor{textcolor}%
\pgfsetfillcolor{textcolor}%
\pgftext[x=2.495104in,y=0.538889in,left,base]{\color{textcolor}\rmfamily\fontsize{15.000000}{18.000000}\selectfont GossipCop}%
\end{pgfscope}%
\end{pgfpicture}%
\makeatother%
\endgroup%

%% file: main.bbl
\begin{thebibliography}{10}
\providecommand{\url}[1]{#1}
\csname url@samestyle\endcsname
\providecommand{\newblock}{\relax}
\providecommand{\bibinfo}[2]{#2}
\providecommand{\BIBentrySTDinterwordspacing}{\spaceskip=0pt\relax}
\providecommand{\BIBentryALTinterwordstretchfactor}{4}
\providecommand{\BIBentryALTinterwordspacing}{\spaceskip=\fontdimen2\font plus
\BIBentryALTinterwordstretchfactor\fontdimen3\font minus
  \fontdimen4\font\relax}
\providecommand{\BIBforeignlanguage}[2]{{%
\expandafter\ifx\csname l@#1\endcsname\relax
\typeout{** WARNING: IEEEtran.bst: No hyphenation pattern has been}%
\typeout{** loaded for the language `#1'. Using the pattern for}%
\typeout{** the default language instead.}%
\else
\language=\csname l@#1\endcsname
\fi
#2}}
\providecommand{\BIBdecl}{\relax}
\BIBdecl

\bibitem{islam2020covid}
M.~S. Islam, T.~Sarkar, S.~H. Khan, A.-H.~M. Kamal, S.~M. Hasan, A.~Kabir,
  D.~Yeasmin, M.~A. Islam, K.~I.~A. Chowdhury, K.~S. Anwar \emph{et~al.},
  ``Covid-19--related infodemic and its impact on public health: A global
  social media analysis,'' \emph{The American journal of tropical medicine and
  hygiene}, vol. 103, no.~4, p. 1621, 2020.

\bibitem{nieves2021infodemic}
G.~M. Nieves-Cuervo, E.~F. Manrique-Hern{\'a}ndez, A.~F. Robledo-Colonia, and
  A.~E.~K. Grillo, ``Infodemic: fake news and covid-19 mortality trends in six
  latin american countriesinfodemia: not{\'\i}cias falsas e tend{\^e}ncias na
  mortalidade por covid-19 em seis pa{\'\i}ses da am{\'e}rica latina,''
  \emph{Revista Panamericana de Salud Publica= Pan American Journal of Public
  Health}, vol.~45, pp. e44--e44, 2021.

\bibitem{shu_fakenewsnet_2018}
K.~Shu, D.~Mahudeswaran, S.~Wang, D.~Lee, and H.~Liu, ``Fakenewsnet: A data
  repository with news content, social context, and spatiotemporal information
  for studying fake news on social media,'' \emph{Big Data}, pp. 171--188,
  2020.

\bibitem{pierri_false_2019}
F.~Pierri and S.~Ceri, ``False news on social media: A data-driven survey,''
  \emph{{SIGMOD} Record}, vol.~48, no.~2, pp. 18--27, 2019.

\bibitem{zhou_fake_2018}
X.~Zhou and R.~Zafarani, ``Fake news: A survey of research, detection methods,
  and opportunities,'' \emph{{arXiv} e-prints}, p. arXiv:1812.00315, 2018.

\bibitem{shu_hierarchical_2019}
K.~Shu, D.~Mahudeswaran, S.~Wang, and H.~Liu, ``Hierarchical propagation
  networks for fake news detection: Investigation and exploitation,'' in
  \emph{Proc. of ICWSM}, 2020.

\bibitem{shu_leveraging_2020}
K.~Shu, G.~Zheng, Y.~Li, S.~Mukherjee, A.~Hassan~Awadallah, S.~Ruston, and
  H.~Liu, ``Leveraging multi-source weak social supervision for early detection
  of fake news,'' \emph{{arXiv} e-prints}, p. arXiv:2004.01732, 2020.

\bibitem{wang_eann_2018}
Y.~Wang, F.~Ma, Z.~Jin, Y.~Yuan, G.~Xun, K.~Jha, L.~Su, and J.~Gao, ``{EANN}:
  Event adversarial neural networks for multi-modal fake news detection,'' in
  \emph{Proc. of SIGKDD}, 2018.

\bibitem{monti_fake_2019}
F.~Monti, F.~Frasca, D.~Eynard, D.~Mannion, and M.~M. Bronstein, ``Fake news
  detection on social media using geometric deep learning,'' \emph{{arXiv}
  e-prints}, p. arXiv:1902.06673, 2019.

\bibitem{shu_defend_2019}
K.~Shu, L.~Cui, S.~Wang, D.~Lee, and H.~Liu, ``{DEFEND}: Explainable fake news
  detection,'' in \emph{Proc. of SIGKDD}, 2019, pp. 395--405.

\bibitem{yang_unsupervised_2019}
S.~Yang, K.~Shu, S.~Wang, R.~Gu, F.~Wu, and H.~Liu, ``Unsupervised fake news
  detection on social media: A generative approach,'' \emph{Proc. of AAAI},
  2019.

\bibitem{hosseinimotlagh2018unsupervised}
S.~Hosseinimotlagh and E.~E. Papalexakis, ``Unsupervised content-based
  identification of fake news articles with tensor decomposition ensembles,''
  in \emph{Proc. of MIS2}, 2018.

\bibitem{gangireddy2020unsupervised}
S.~C.~R. Gangireddy, C.~Long, and T.~Chakraborty, ``Unsupervised fake news
  detection: A graph-based approach,'' in \emph{Proc. of HyperText}, 2020.

\bibitem{li2021unsupervised}
D.~Li, H.~Guo, Z.~Wang, and Z.~Zheng, ``Unsupervised fake news detection based
  on autoencoder,'' \emph{IEEE Access}, vol.~9, pp. 29\,356--29\,365, 2021.

\bibitem{silva2021embracing}
A.~Silva, L.~Luo, S.~Karunasekera, and C.~Leckie, ``Embracing domain
  differences in fake news: Cross-domain fake news detection using multimodal
  data,'' in \emph{Proc. of AAAI}, 2021.

\bibitem{cui2020coaid}
L.~Cui and D.~Lee, ``Coaid: Covid-19 healthcare misinformation dataset,''
  \emph{{arXiv} e-prints}, p. arXiv:2006.00885, 2020.

\bibitem{zhou2020recovery}
X.~Zhou, A.~Mulay, E.~Ferrara, and R.~Zafarani, ``Recovery: A multimodal
  repository for covid-19 news credibility research,'' in \emph{Proc. of CIKM},
  2020.

\bibitem{zhou2021hidden}
X.~Zhou, H.~Elfardy, C.~Christodoulopoulos, T.~Butler, and M.~Bansal, ``Hidden
  biases in unreliable news detection datasets,'' in \emph{Proc. of EACL},
  2021.

\bibitem{nakov2020can}
P.~Nakov, ``Can we spot the" fake news" before it was even written?''
  \emph{arXiv preprint arXiv:2008.04374}, 2020.

\bibitem{pehlivanoglu2021role}
D.~Pehlivanoglu, T.~Lin, F.~Deceus, A.~Heemskerk, N.~C. Ebner, and B.~S.
  Cahill, ``The role of analytical reasoning and source credibility on the
  evaluation of real and fake full-length news articles,'' \emph{Cognitive
  research: principles and implications}, vol.~6, no.~1, pp. 1--12, 2021.

\bibitem{horne2019different}
B.~D. Horne, J.~N{\o}rregaard, and S.~Adal{\i}, ``Different spirals of
  sameness: A study of content sharing in mainstream and alternative media,''
  in \emph{Proc. of ICWSM}, 2019.

\bibitem{horne2018exploration}
B.~D. Horne and S.~Adali, ``An exploration of verbatim content republishing by
  news producers,'' in \emph{Proc. of NECO}, 2018.

\bibitem{baly:2018:EMNLP2018}
R.~Baly, G.~Karadzhov, D.~Alexandrov, J.~Glass, and P.~Nakov, ``Predicting
  factuality of reporting and bias of news media sources,'' in \emph{Proc. of
  EMNLP}, 2018.

\bibitem{baly:2020:ACL2020}
R.~Baly, G.~Karadzhov, J.~An, H.~Kwak, Y.~Dinkov, A.~Ali, J.~Glass, and
  P.~Nakov, ``What was written vs. who read it: News media profiling using text
  analysis and social media context,'' in \emph{Proc. of ACL}, 2020.

\bibitem{silva2020meteor}
A.~Silva, S.~Karunasekera, C.~Leckie, and L.~Luo, ``{METEOR: Learning Memory
  and Time Efficient Representations from Multi-modal Data Streams},'' in
  \emph{Proc. of CIKM}, 2020.

\bibitem{silva2020omba}
A.~Silva, L.~Luo, S.~Karunasekera, and C.~Leckie, ``{OMBA: User-Guided Product
  Representations for Online Market Basket Analysis},'' in \emph{Proc. of
  ECML-PKDD}, 2020.

\bibitem{zhang2017react}
C.~Zhang, K.~Zhang, Q.~Yuan, F.~Tao, L.~Zhang, T.~Hanratty, and J.~Han,
  ``React: Online multimodal embedding for recency-aware spatiotemporal
  activity modeling,'' in \emph{Proc. of SIGIR}, 2017.

\bibitem{maaten2008visualizing}
L.~v.~d. Maaten and G.~Hinton, ``Visualizing data using t-sne,'' \emph{Journal
  of Machine Learning Research}, vol.~9, pp. 2579--2605, 2008.

\bibitem{liu2019roberta}
Y.~Liu, M.~Ott, N.~Goyal, J.~Du, M.~Joshi, D.~Chen, O.~Levy, M.~Lewis,
  L.~Zettlemoyer, and V.~Stoyanov, ``Roberta: A robustly optimized bert
  pretraining approach,'' \emph{{arXiv} e-prints}, p. arXiv:1907.11692, 2019.

\bibitem{mohammad2013nrc}
S.~M. Mohammad and P.~D. Turney, ``Nrc emotion lexicon,'' \emph{National
  Research Council, Canada}, vol.~2, 2013.

\bibitem{pennebaker_development_2015}
\BIBentryALTinterwordspacing
J.~W. Pennebaker, R.~L. Boyd, K.~Jordan, and K.~Blackburn, ``The development
  and psychometric properties of {LIWC}2015,'' Tech. Rep., 2015. [Online].
  Available: \url{https://repositories.lib.utexas.edu/handle/2152/31333}
\BIBentrySTDinterwordspacing

\bibitem{mc1969smog}
G.~H. Mc~Laughlin, ``Smog grading-a new readability formula,'' \emph{Journal of
  reading}, vol.~12, no.~8, pp. 639--646, 1969.

\bibitem{graham2009liberals}
J.~Graham, J.~Haidt, and B.~A. Nosek, ``Liberals and conservatives rely on
  different sets of moral foundations.'' \emph{Journal of personality and
  social psychology}, vol.~96, no.~5, p. 1029, 2009.

\bibitem{chakraborty2016stop}
A.~Chakraborty, B.~Paranjape, S.~Kakarla, and N.~Ganguly, ``Stop clickbait:
  Detecting and preventing clickbaits in online news media,'' in \emph{Proc. of
  ASONAM}, 2016.

\bibitem{dahal2018learning}
P.~Dahal, ``Learning embedding space for clustering from deep
  representations,'' in \emph{Proc. of IEEE BigData}, 2018.

\bibitem{eldele2021time}
E.~Eldele, M.~Ragab, Z.~Chen, M.~Wu, C.~K. Kwoh, X.~Li, and C.~Guan,
  ``Time-series representation learning via temporal and contextual
  contrasting,'' in \emph{Proc. of IJCAI}, 2021.

\bibitem{silva2021propagation2vec}
A.~Silva, Y.~Han, L.~Luo, S.~Karunasekera, and C.~Leckie, ``Propagation2vec:
  Embedding partial propagation networks for explainable fake news early
  detection,'' \emph{Information Processing \& Management}, vol.~58, no.~5, p.
  102618, 2021.

\bibitem{silva-embedding-2020}
------, ``Embedding partial propagation network for fake news early
  detection.'' in \emph{Proc. of MAISON}, 2020.

\bibitem{velivckovic2017graph}
P.~Veli{\v{c}}kovi{\'c}, G.~Cucurull, A.~Casanova, A.~Romero, P.~Lio, and
  Y.~Bengio, ``Graph attention networks,'' \emph{arXiv preprint
  arXiv:1710.10903}, 2017.

\bibitem{velickovic2019deep}
P.~Velickovic, W.~Fedus, W.~L. Hamilton, P.~Li{\`o}, Y.~Bengio, and R.~D.
  Hjelm, ``Deep graph infomax,'' in \emph{Proc. of ICLR}, 2019.

\bibitem{ovalle2017gated}
J.~E.~A. Ovalle, T.~Solorio, M.~Montes-y G{\'o}mez, and F.~A. Gonz{\'a}lez,
  ``Gated multimodal units for information fusion.'' in \emph{Proc. of ICLR},
  2017.

\bibitem{baevski2022data2vec}
A.~Baevski, W.-N. Hsu, Q.~Xu, A.~Babu, J.~Gu, and M.~Auli, ``Data2vec: A
  general framework for self-supervised learning in speech, vision and
  language,'' \emph{arXiv preprint arXiv:2202.03555}, 2022.

\bibitem{grill2020bootstrap}
J.-B. Grill, F.~Strub, F.~Altch{\'e}, C.~Tallec, P.~Richemond, E.~Buchatskaya,
  C.~Doersch, B.~Avila~Pires, Z.~Guo, M.~Gheshlaghi~Azar \emph{et~al.},
  ``Bootstrap your own latent-a new approach to self-supervised learning,''
  \emph{Advances in Neural Information Processing Systems}, vol.~33, pp.
  21\,271--21\,284, 2020.

\bibitem{liu2020peer}
Y.~Liu and H.~Guo, ``Peer loss functions: Learning from noisy labels without
  knowing noise rates,'' in \emph{Proc. of ICML}, 2020.

\bibitem{berthon2021confidence}
A.~Berthon, B.~Han, G.~Niu, T.~Liu, and M.~Sugiyama, ``Confidence scores make
  instance-dependent label-noise learning possible,'' in \emph{Proc. of ICML},
  2021.

\bibitem{silva2022noise}
A.~Silva, L.~Luo, S.~Karunasekera, and C.~Leckie, ``Noise-robust learning from
  multiple unsupervised sources of inferred labels,'' in \emph{Proc. of AAAI},
  2022.

\bibitem{li2021contrastive}
Y.~Li, P.~Hu, Z.~Liu, D.~Peng, J.~T. Zhou, and X.~Peng, ``Contrastive
  clustering,'' in \emph{Proc. of AAAI}, 2021.

\bibitem{chen2020simple}
T.~Chen, S.~Kornblith, M.~Norouzi, and G.~Hinton, ``A simple framework for
  contrastive learning of visual representations,'' in \emph{Proc. of ICML},
  2020.

\bibitem{chuang2022robust}
C.-Y. Chuang, R.~D. Hjelm, X.~Wang, V.~Vineet, N.~Joshi, A.~Torralba,
  S.~Jegelka, and Y.~Song, ``Robust contrastive learning against noisy views,''
  \emph{arXiv preprint arXiv:2201.04309}, 2022.

\bibitem{yin2008truth}
X.~Yin, J.~Han, and S.~Y. Philip, ``Truth discovery with multiple conflicting
  information providers on the web,'' \emph{IEEE transactions on knowledge and
  data engineering}, 2008.

\bibitem{yang2019unsupervised}
S.~Yang, K.~Shu, S.~Wang, R.~Gu, F.~Wu, and H.~Liu, ``Unsupervised fake news
  detection on social media: A generative approach,'' in \emph{Proc. of AAAI},
  2019.

\bibitem{bruff2005assignment}
D.~Bruff, ``The assignment problem and the hungarian method,'' \emph{Notes for
  Math}, 2005.

\bibitem{yang_hierarchical_2016}
Z.~Yang, D.~Yang, C.~Dyer, X.~He, A.~Smola, and E.~Hovy, ``Hierarchical
  attention networks for document classification,'' in \emph{Proc. of
  NAACL-HLT}, 2016.

\bibitem{perez-rosas_automatic_2018}
V.~Pérez-Rosas, B.~Kleinberg, A.~Lefevre, and R.~Mihalcea, ``Automatic
  detection of fake news,'' in \emph{Proc. of COLING}, 2018.

\bibitem{wu_tracing_2018}
L.~Wu and H.~Liu, ``Tracing fake-news footprints: Characterizing social media
  messages by how they propagate,'' in \emph{Proc. of WSDM}, 2018.

\bibitem{ma_detect_2017}
J.~Ma, W.~Gao, and K.-F. Wong, ``Detect rumors in microblog posts using
  propagation structure via kernel learning,'' in \emph{Proc. of ACL}, 2017.

\end{thebibliography}
